\documentclass[12pt]{clear2027}
\makeatletter
\renewcommand\ps@jmlrtps{\let\@mkboth\@gobbletwo\def\@oddhead{}\let\@evenhead\@oddhead
  \def\@oddfoot{}\let\@evenfoot\@oddfoot}
\renewcommand\ps@jmlrps{\let\@mkboth\@gobbletwo\def\@oddhead{}\let\@evenhead\@oddhead
  \def\@oddfoot{\hfill\small\rmfamily\thepage\hfill}\let\@evenfoot\@oddfoot}
\makeatother

\usepackage[T1]{fontenc}
\usepackage{times}
\usepackage{booktabs}
\usepackage{longtable}

\usepackage{pdflscape}
\usepackage{array}
\newcolumntype{L}[1]{>{\raggedright\arraybackslash}p{#1}}
\usepackage{xcolor}
\usepackage{colortbl}
\definecolor{shade}{HTML}{3C7FA8}
\definecolor{shadetan}{HTML}{B07A3E}
\definecolor{shadepurple}{HTML}{8A5BAA}
\usepackage{enumitem}
\setlist{nosep, leftmargin=1.5em}
\usepackage{tikz}
\usetikzlibrary{arrows.meta,positioning}

\input{generated/numbers.tex}
\input{generated/numbers_selection.tex}
\input{generated/numbers_lists.tex}
\input{generated/numbers_errors.tex}
\input{generated/numbers_overall.tex}
\input{generated/numbers_separability.tex}
\newcommand{\SortPlants}{1,190}

\newcommand{\SortVarFp}{9\%}
\newcommand{\SortRtwoFp}{6\%}
\newcommand{\SortRandFp}{13\%}

\newcommand{\ConfPlants}{10}
\newcommand{\SibRef}{0.10}
\newcommand{\SibFloor}{0.06}
\newcommand{\SibPlants}{10}

\newcommand{\Y}{Y}
\newcommand{\pa}{\mathrm{Pa}}

\title[Target-Centric Causal Discovery in Industrial Processes]{%
  A Synthetic Benchmark for Target-Centric Causal Discovery\\ in Industrial Processes}

\clearauthor{\Name{Angelo Ziletti} \Email{angelo@svelaro.com}\\ \addr Svelaro GmbH}

\begin{document}

\maketitle

\begin{abstract}%
Causal discovery promises to find which recorded variables drive a process target such as product quality, yield or
energy use. Process data with known causes are rare, and in existing benchmarks the difficulty of identifying the
causes of a target is either fixed by the recorded data or set for the whole graph. We present a synthetic benchmark for target-centric
causal discovery in industrial processes. Its generator is a two-level structural causal model inspired by batch
production, assembled from components that each model one phenomenon of process data, such as a control loop or a
delayed effect. The graph is built around the target: without sensor error, every non-cause can be distinguished from the
direct causes by conditioning on them or by time order. Each component is set independently, so the difficulty can be varied one condition at a time. We evaluate eight methods, from association measures to effect estimation,
with an error analysis by the role of the selected candidate; all degrade as the
difficulties combine, each in a way consistent with its assumptions.%
\end{abstract}

\begin{keywords}%
causal discovery; time series; benchmark; root-cause analysis; batch processes; structural causal models%
\end{keywords}

\section{Introduction}
\label{sec:intro}

Many industrial processes run as repeated episodes. In batch production, a vessel is filled, heated, held while the
reaction runs, cooled and emptied, and the same recipe is run hundreds of times; in every batch, tens of process
variables are logged minute by minute, together with a target of interest, such as the product quality, the yield or the energy
use of the batch. After an off-target batch, the engineer needs the recorded variables that act on the target
directly, since a variable merely correlated with the target is no lever for intervention. This is local causal discovery
\citep{aliferis2010local,yu2020featureselection,ling2025hlcd} on time series observed over many episodes; root cause analysis is a main application of causal discovery in
manufacturing \citep{vukovic2022manufacturing}. Causal discovery from
time series, however, is usually evaluated on the graph over all variables; how well a method finds the direct
causes of one target is a different question, and current benchmarks are not built to answer it.

Benchmarks built on real data or fitted to it
\citep{cheng2024causaltime,stein2025causalrivers} are realistic, but few plants release data, and those from
industrial processes \citep{menegozzo2022cipcad,mogensen2024ess,mehling2026causrca} come with expert graphs, some only between subsystems, whereas simulated benchmarks
\citep{herdeanu2025causaldynamics,gobler2024causalassembly} know their graph exactly. Any of them allows a target to be chosen and its
direct causes scored, but the conditions that make those causes hard to find, how strong they are, what disturbs
them and whether a controller acts on them, are fixed in the recorded datasets and, in the generators, set for the
whole graph instead of for the chosen target.

We present a synthetic benchmark built around the target. Its generator is a structural causal model inspired by
batch production, assembled from components; each is a causal mechanism whose effect on the task is known and is set
by one parameter, so the difficulty of finding the causes of the target can be varied one condition at a time
(Section~\ref{sec:generator}). Our contributions are (i) a \emph{target-centric benchmark}: the task, the roles of the
candidates (direct cause, ancestor, sibling or descendant) and the difficulty are all defined around one target
(Sections~\ref{sec:problem} and~\ref{sec:setup}); (ii) a \emph{configurable generator} of batch plants
(Section~\ref{sec:generator}); and (iii) an \emph{evaluation} of eight methods, from association measures to
effect estimation, with an error analysis (Sections~\ref{sec:protocol} to~\ref{sec:experiments}).

The generator, the configurations with their seeds and a sample of the data are
released.\footnote{\url{https://github.com/angeloziletti/svelaro-industrial-causal-generator}}

\section{Related work}
\label{sec:related}

Surveys review temporal causal discovery \citep{gong2024temporal,runge2019earth}; \citet{assaad2022survey} also
evaluate methods on simulated systems and a few real datasets. CausalTime fits neural generative models to real series and derives the graph from the fitted model
\citep{cheng2024causaltime}; CausalRivers provides river-discharge networks with graphs built from metadata
\citep{stein2025causalrivers}; CausalDynamics \citep{herdeanu2025causaldynamics} generates coupled differential equations with noise, confounding
and lags; causal chambers are a physical testbed with controllable inputs
\citep{gamella2025chambers}; CausalProfiler samples tabular benchmarks from explicit assumptions about models, queries
and data \citep{panayiotou2026causalprofiler}; causalAssembly generates semi-synthetic per-part data of an assembly line from
a graph set by engineers and mechanisms fitted to recorded data, scored on the full graph
\citep{gobler2024causalassembly}. \citet{lawrence2020dgp} argue, as we do, for a data-generating process with
explicit knobs. For process data, Tennessee
Eastman \citep{downs1993te,reinartz2021te} and IndPenSim \citep{goldrick2015indpensim,goldrick2019indpensim} are the
standard simulated continuous and fed-batch processes, CIPCaD-Bench pairs data simulated from the former with real
data from a food-processing line \citep{menegozzo2022cipcad}, the cryoplant of the European Spallation Source
provides real plant data with an expert graph between subsystems \citep{mogensen2024ess}, causRCA provides lathe
recordings with an expert graph for root-cause analysis \citep{mehling2026causrca}, and root-cause benchmarks exist
for software systems \citep{hardt2024petshop}; each provides a fixed set of systems and graphs. Our benchmark differs
in three respects: the task is defined for one target, difficulty is set through variance-based controls
relative to that target, and every candidate of every plant carries its role with respect to that target, in addition
to the full graph.

\section{The task}
\label{sec:problem}

A process is observed over $B$ episodes, repeated runs of the same process such as the batches of a batch plant. In
each episode, $p$ variables $X_1,\dots,X_p$ and a target $\Y$ are recorded at $T$ time steps; the other variables of
the process are latent, as a plant logs its process variables but not every condition that acts on them. The task is to find the \emph{direct causes} $D$ of $\Y$ among the candidates $\mathcal{C}$:
the candidates with an edge into $\Y$ in the latent projection of the causal graph onto the recorded variables, in
which an edge joins two variables when one acts on the other directly or only through latent variables. Their effect
on $\Y$ may be delayed and nonlinear. A method returns a ranking of all candidates and a selected list, which may be
empty.

Each candidate has one role for the target $\Y$, read on the generating graph $\mathcal{G}$ (Section~\ref{sec:scm}).
It is a \emph{direct cause}; an \emph{ancestor}, which influences $\Y$ only through other recorded variables; a
\emph{descendant}, which $\Y$ influences; or a \emph{sibling}, which shares a latent cause with a direct cause and is
correlated with $\Y$ without affecting it. The roles partition the candidates by their ancestral relation to $\Y$,
which the latent projection preserves, so they are defined by $\mathcal{G}$ and the set of recorded variables, as
direct causation is in any causal graph over a variable set \citep{spirtes2000causation}. In our generator, every
candidate that is not a direct cause also shares a latent cause with $\Y$, so every wrong candidate is associated
with $\Y$. We count only the direct causes as correct
answers; ancestors, although they influence $\Y$, count as wrong in our evaluation. We release the role of every
candidate, so that other scorings are possible. Without sensor error, every wrong candidate is either separated from
the target given its direct causes or distinguished by time order (Appendix~\ref{app:identifiability}). These labels
describe the generating graph; they do not guarantee recovery of the direct causes from observational data.

\section{The generator}
\label{sec:generator}

\begin{figure}[t]
\centering
\includegraphics[width=\linewidth]{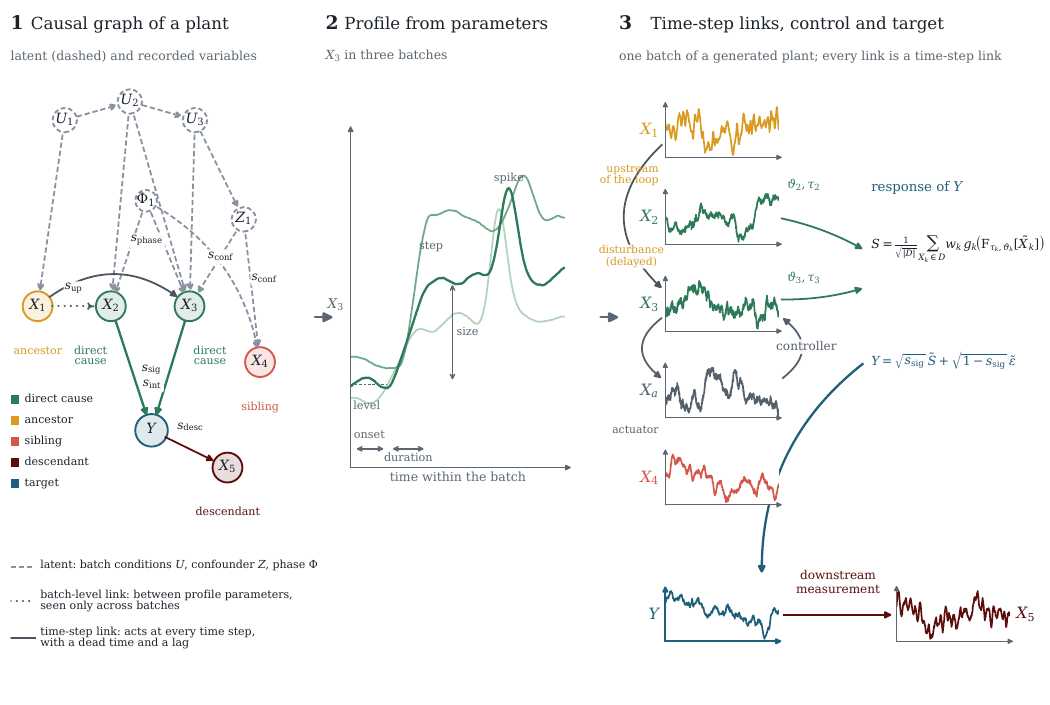}
\caption{How the generator produces data. (1) The causal graph of a plant and the role of each recorded variable
for $\Y$. (2) The profile of $X_3$ in three batches, built from its profile parameters $\theta(X_3)$. (3) One batch:
time-step links, control and the target.}
\label{fig:layers}
\end{figure}

The generator is a structural causal model (SCM) whose variables are defined at two levels, one value per batch and one
value per time step, and whose mechanisms are built from components inspired by batch production
(Table~\ref{tab:components}, Figure~\ref{fig:layers}). To fix ideas, we illustrate the components with a
generic batch process: raw materials are charged into a vessel, transformed under controlled conditions such as
temperature and mixing, and the product is then separated and dried downstream; the target is an outcome of the batch,
e.g.\ the quality of the product, the yield or the energy use.

\begin{table}[t]
\centering\scriptsize
\caption{Core components of the generator.}
\label{tab:components}
\label{tab:variables}
\label{tab:factory}
\label{tab:axes}
\begin{tabular}{@{}L{0.235\linewidth} L{0.355\linewidth} L{0.355\linewidth}@{}}
\toprule
Component & In the generator & In a batch plant \\
\midrule
\multicolumn{3}{@{}l}{\emph{Once per batch}} \\
batch profile $\theta(X_i)$ & 2--4 motifs per variable (step, ramp, spike, plateau); their onsets, durations, sizes and the level form $\theta(X_i)$ & a temperature or flow rises, holds and falls in every batch, at different times and by different amounts \\ \addlinespace[2pt]
batch conditions $U_j$ & act on the profiles of many variables & conditions that differ between batches and affect many sensors, e.g.\ coolant or outside temperature \\ \addlinespace[2pt]
confounders $Z_k$, $Z_k(t)$ & move a direct cause and its siblings together, also within the batch & an unmeasured property of a raw material, e.g.\ its purity, that moves several sensors together \\ \addlinespace[2pt]
batch-level links & between profile parameters, without lag; seen only across batches & one batch quantity sets another, e.g.\ a larger charge gives a higher temperature peak \\
\midrule
\multicolumn{3}{@{}l}{\emph{At every time step within a batch}} \\
time-step links $\mathrm{F}_{\tau,\vartheta}$ & first-order response with a dead time & a change propagates to another variable after a delay, e.g.\ the pressure in the vessel follows its temperature \\ \addlinespace[2pt]
controller $\kappa$ & holds a direct cause by moving an actuator (not a candidate) & a control loop moves a valve, e.g.\ the coolant valve to hold the vessel temperature \\ \addlinespace[2pt]
target response $g_k$, $s_{\mathrm{int}}$ & linear, saturating, optimal or threshold; joint effects of pairs of causes & quality improves and then levels off, is best at one temperature, or drops past an impurity limit; some causes matter only together, e.g.\ temperature and the amount of a raw material \\ \addlinespace[2pt]
unexplained part $s_{\mathrm{sig}}$ & a slow part of $\Y$ no recorded variable explains & causes that are not measured \\ \addlinespace[2pt]
downstream measurements & respond to $\Y$ with a delay & measured after the product is formed, e.g.\ a pressure in the separation step or the moisture after drying \\
\bottomrule
\end{tabular}
\end{table}

\subsection{A two-level structural causal model}
\label{sec:scm}

The generator draws an SCM, a causal graph $\mathcal{G}$ with an assignment for each of its nodes, which is then fixed;
we call one such SCM a \emph{plant}. Its structure comprises the motifs of every recorded variable, the graph and its
weights, and the direct causes of $\Y$ with their response functions. Its data are $B$ batches, each a sample of the SCM
with newly sampled exogenous noise.

A batch plant produces data on two time scales. Each batch is run from a recipe and leaves a batch record, which
lists, once per batch, the amounts charged, the lots of the raw materials, and the setpoints and duration of each
phase. During the batch, sensors log trajectories at a fixed interval. The recipe prescribes the same
sequence of phases for every batch, e.g.\ a heat-up, a hold and a cool-down, so the trajectories recur from batch to
batch; they differ in when each part starts, how long it lasts, how high it rises and sometimes whether it occurs at
all. Causal effects act on both time scales. Some act from one moment to the next:
the pressure in the vessel follows its temperature within minutes, in every batch. Others act between properties of
the whole batch, e.g.\ a larger charge can give a higher temperature peak; these appear most clearly when batches
are compared, and batch data are commonly analysed by comparing batches \citep{nomikos1994multiway}.

The generator reproduces this structure with an SCM whose variables are defined at two levels. \emph{Batch-level}
variables take one value per batch: the latent batch conditions $U_j$, the confounders $Z_k$, the phase timings
$\Phi_q$, and the profile parameters $\theta(X_i)$ of every recorded variable, i.e.\ the onset, duration and
size of each motif of its profile and its level. \emph{Time-step} variables take one value per time step
${t = 1,\dots,T}$ of the batch: the recorded variables $X_i(t)$, the target $\Y(t)$, the actuator of the controller and
the within-batch part $Z_k(t)$ of each confounder.

The generator follows the order of a plant: in each batch, the batch level is drawn first, with new noise. A batch is
generated in five steps (Figure~\ref{fig:layers}): (i) the batch conditions and confounders are drawn; (ii) they set the profile
parameters, each of which sets the timing or size of one motif of its own variable, e.g.\ the start of the heating
ramp; (iii) the profiles are built from these parameters and the within-batch part of each confounder is added; (iv) the time-step links and the controller act within the
batch; (v) the target is computed from its direct causes, followed by the downstream measurements.

The two levels give two kinds of causal links between recorded variables. A \emph{batch-level link} acts when the
profiles are set: a parameter of one variable, e.g.\ the size of the charging step of the feed, is a parent of a
parameter of another, e.g.\ the height of the temperature peak. Each batch gives a single value of each parameter, so
the link appears only across batches. A \emph{time-step link} acts within the
batch: the effect follows the cause after a dead time $\vartheta$, in every batch, and the order in time shows the
direction.

The two kinds of links leave different evidence. $\Y$ responds to its direct causes through
time-step links, which can be read from the time series. The siblings, and the candidates linked to a direct cause at
the batch level, are instead tied to $\Y$ through shared batch conditions, confounders and batch-level links, and
these ties appear mostly when batches are compared. Methods that pool many batches see both.

The graph $\mathcal{G}$, unrolled over the time steps of a batch, is acyclic: the controller and the
operator reactions act with a delay of at least one step, so their feedback loops become chains in time, and one
batch can be generated by evaluating the nodes in that order (Algorithm~\ref{alg:episode}). Batches are
independent draws of the SCM unless the additional lots and drift are switched on; these add batch-level variables
linked from one batch to the next (Section~\ref{sec:optional}). A recorded variable is the group of its nodes,
$\{\theta(X_i)\} \cup \{X_i(t)\}_{t}$, and its role for $\Y$ (Section~\ref{sec:problem}) is read on $\mathcal{G}$;
the methods receive only the trajectories $X_i(t)$ and $\Y(t)$. Every plant is released with its data, candidate roles,
separation or time-order labels for non-causes, the unique share of every direct cause and its full graph $\mathcal{G}$
(Appendix~\ref{app:fullgraph}).

\subsection{Mechanisms}
\label{sec:mechanisms}

\paragraph{Batch level.}\label{sec:episodescm} \emph{Batch conditions.} Each plant has $m$ latent batch conditions
$U_1,\dots,U_m$ (e.g.\ the temperature of the coolant or of the surroundings), linked by a random directed tree
in which each condition acts on one other.

\emph{Confounders.} Each direct cause $X_k$ shares a latent confounder $Z_k$ (e.g.\ the purity of a raw material),
drawn independently in each batch, with the non-causes assigned to it (every non-cause is assigned to one direct
cause). It acts in two stages. Once per batch, $Z_k$ enters their profile parameters, so the motifs of a cause and
its siblings start, last and rise together. Its slow component $Z_k(t)$ is then added to their trajectories, so they
also move together from one time step to the next. $Z_k$ reaches $\Y$ only through $X_k$. The non-causes that reach $\Y$ by no other route, the \emph{siblings},
are therefore confounded with $\Y$: they move with $X_k$ without affecting $\Y$.

\emph{Profile parameters.} Each parameter $\theta$ is the mean of its parents plus Gaussian noise, mapped into its
admissible range (Equation~\ref{eq:param}, Appendix~\ref{app:optional}). Its parents are batch conditions,
confounders, phase timings (Section~\ref{sec:optional}) and parameters of other recorded variables; the weights of the
confounders, phases and batch-level links are instead set by their shares (Section~\ref{sec:shares}).

\paragraph{Time-step level.}\label{sec:motifs}\label{sec:coupling} \emph{Profiles.} Each recorded variable $X_i$
follows a profile that recurs in every batch, built from a small library of \emph{motifs}: a step, a ramp, a spike
and a plateau. A variable has two to four motifs, each up or down; their shapes, directions and order are drawn once
per plant and then fixed. The batch is divided into consecutive slots, one per motif, and in each batch the motif
starts at an onset within its slot, lasts a duration and has a size; these, with the level, are the profile
parameters. The profile is the level plus the sum of the motifs, plus a smooth random texture and white noise
(Appendix~\ref{app:params}). Other families of profiles could
replace the motifs without changing the rest of the generator, e.g.\ compositions of Gaussian-process kernels as in
CauKer \citep{xie2026cauker}.

\emph{Time-step links.} Within a batch, a variable responds to its upstream variables with a dead time and a lag, e.g.\
the pressure in the vessel follows its temperature. We use the first-order-plus-dead-time response that is
standard in process control: the input is delayed by the dead time $\vartheta$ and then smoothed exponentially with
time constant $\tau$, $\mathrm{F}_{\tau,\vartheta}[x](t) = y(t)$ with $y(t) = y(t-1) + \big(x(t - \vartheta) -
y(t-1)\big)/\tau$. A variable with upstream variables receives the sum of their delayed responses, each with its own gain, mixed
with its own profile in the proportion set by the upstream share $s_{\mathrm{up}}$. Recorded variables are ordered
(upstream variables, direct causes, other variables), and all links, batch-level and time-step, point forward in this
order; a variable may carry links of both kinds. This gives $\Y$ ancestors that act only through its direct causes, and non-causes that follow the direct
causes.

\emph{Controller.} The main variables of a plant are under feedback control, which holds them near a setpoint and
hides their variation. The generator puts one direct cause $X_c$, e.g.\ the vessel temperature, under control: a
recorded actuator $X_a$, e.g.\ the coolant valve, is moved by a control rule so that $X_c$ follows its setpoint, the
trajectory the recipe prescribes over the batch (the profile of $X_c$). A recorded upstream variable $X_d$, the
\emph{disturbance of the controlled cause}, e.g.\ a colder raw material entering the vessel, pushes $X_c$ away from
it. The rule is integral, with tightness $\kappa \in [0,1]$: its correction accumulates the gap between setpoint and
$X_c$ and is added to the actuator's own profile (Equation~\ref{eq:loop}). At ${\kappa = 0}$ the controller is off and
$X_d$ is an ordinary ancestor of $\Y$, scored as wrong like any ancestor. As $\kappa$ grows, the slow part of the
disturbance moves from $X_c$ to the actuator: $X_c$ still causes $\Y$ but varies less, so its effect is harder to
detect, and $X_d$ remains a cause of $\Y$, but the loop cancels its effect, so the data may show no dependence; with
the controller on, it is therefore left out of the candidates. The actuator, a
controller output known from the plant's diagram, is never a candidate.

\paragraph{Target and downstream measurements.}\label{sec:target} Each direct cause $X_k$ acts on $\Y$ through a
delayed response followed by a response function drawn per cause from a library of four,
\begin{equation}
\label{eq:target}
S(t) = \frac{1}{\sqrt{|D|}} \sum_{X_k \in D} w_k\, g_k\!\left(\mathrm{F}_{\tau_k,\vartheta_k}[\tilde X_k](t)\right),
\qquad g_k \in \{\,x,\ \tanh x,\ x^2 - 1,\ \max(0, x - 1)\,\},
\end{equation}
where ${\tilde z = (z-\mu_z)/\sigma_z}$ uses reference mean $\mu_z$ and standard deviation $\sigma_z$, fixed under
intervention, and $w_k$ is the weight of the cause. The library holds the
simplest form of each way a process variable can act on quality: proportional, saturating, optimum-shaped (best at
one value and worse on either side, so that the cause is nearly uncorrelated with the target) and thresholded (no
effect until a tolerance is exceeded); other forms can be added. $S$ can also contain interaction terms, products of two direct
causes, e.g.\ the vessel temperature acting only together with the amount of a raw material; the interaction share of
Section~\ref{sec:shares} sets how much of $S$ they carry. The target mixes this signal with a part
$\varepsilon$ that no recorded variable explains,
\begin{equation}
\label{eq:y}
\Y(t) = \sqrt{s_{\mathrm{sig}}}\,\tilde S(t) + \sqrt{1 - s_{\mathrm{sig}}}\,\tilde\varepsilon(t),
\end{equation}
where $\tilde S$ and $\tilde\varepsilon$ are the standardised signal and noise, the signal share
$s_{\mathrm{sig}}$ sets the part of $\Y$ explained by its direct causes (Section~\ref{sec:shares}), and $\varepsilon$ varies on the time
scale of $S$. Downstream measurements influence no direct cause; each receives
the delayed response $\mathrm{F}_{\tau,\vartheta}[\Y]$, mixed with its own profile in the proportion set by the
downstream share $s_{\mathrm{desc}}$.

\subsection{Variance-based difficulty controls}
\label{sec:shares}

The generator controls route strength through variance-based weights $s \in [0,1]$, called \emph{shares} below.
For trajectory mixtures, $\tilde a$ and $\tilde r$ denote the standardised route contribution $a$ and remainder $r$,
mixed as
\begin{equation}
\label{eq:share}
\text{variable} = \sqrt{s}\,\tilde a + \sqrt{1-s}\,\tilde r .
\end{equation}
For uncorrelated unit-variance contributions, $s$ equals the route's variance fraction; otherwise it specifies their
relative weighting.

The shares are set along the process, from the inputs of a batch to the downstream measurements
(Figure~\ref{fig:layers}, panel 1). A latent
confounder, e.g.\ the purity of a raw material, moves a direct cause together with its siblings: the
\emph{confounder share} $s_{\mathrm{conf}}$ sets its share of their profile parameters and, relative to the reference,
the strength of its within-batch component (Appendix~\ref{app:realised}). The \emph{upstream share} $s_{\mathrm{up}}$ sets the part of a variable,
e.g.\ a direct cause, explained by its recorded parents. The \emph{signal share} $s_{\mathrm{sig}}$ weights the
causes' contribution to $\Y$; the \emph{interaction share} $s_{\mathrm{int}}$ weights products of pairs of causes
within that contribution. Finally, $\Y$ drives the downstream measurements with the \emph{downstream share}
$s_{\mathrm{desc}}$. Only $s_{\mathrm{sig}}$ and $s_{\mathrm{int}}$ shape $\Y$; the others set how strongly siblings,
ancestors and descendants are associated with $\Y$. In a linear model with independent errors and a single
connecting path, correlations multiply along that path. Additional paths, temporal filtering and nonlinearities
affect the observed associations. Section~\ref{sec:setup} varies these shares, the controller tightness $\kappa$ and
the parameters below, individually or in combination; measured on the generated plants, each share matches its
setting (Table~\ref{tab:dataset}; Appendix~\ref{app:realised}).

\subsection{Additional components}
\label{sec:optional}
\label{sec:batchphen}
\label{sec:imperfections}

The remaining components add difficulties typical of batch plants; Appendix~\ref{app:optional} gives their
equations. Common causes across batches:
\begin{itemize}
  \item \emph{Recipe phases} ($s_{\mathrm{phase}}$): a latent timing per phase window, drawn once per batch,
  shifts the onset of every motif in the window, so all variables shift together at the phase boundaries.
  \item \emph{Raw-material lots} and \emph{campaign drift} ($s_{\mathrm{lot}}$, $s_{\mathrm{drift}}$): a latent lot
  process over batches and a linear trend shift the batch mean of every variable, so batches become dependent and
  every pair of variables shares these common causes.
\end{itemize}
\emph{Imperfections}, which change how the relations are observed or how they act:
\begin{itemize}
  \item \emph{Sensor error} ($s_{\mathrm{err}}$): each reading mixes the true value with independent, slowly drifting
  error, weighted by $s_{\mathrm{err}}$. A method conditions on the reading of a cause, not on its true value,
  so the path from a sibling through the cause to $\Y$ is no longer fully blocked.
  \item \emph{Product grades} ($c$): a latent grade per batch scales the effect of each direct cause in
  proportion $1 + c$ to $1 - c$ between the two grades, so at ${c = 1}$ each cause acts in one grade only.
  \item \emph{Variable delays} ($j$): the dead time of each direct cause grows in each batch by a random amount up to
  $j$ times its time scale, as mixing takes longer in a fuller vessel.
  \item \emph{Cumulative effects} ($k_{\mathrm{avg}}$): $k_{\mathrm{avg}}$ weights each cause's recent average
  against its current response, as when the product depends on the heat received over the last
  hour.
  \item \emph{Operator reactions} ($s_{\mathrm{op}}$): after a deviation of $\Y$, the reference value of one direct cause is moved
  to counter it, one decorrelation time of $\Y$ later, so the graph unrolled over time stays acyclic.
\end{itemize}

\section{Experimental setup}
\label{sec:setup}
\label{sec:protocol}

\paragraph{Plants and configurations.} At the reference, each plant has ${p = 20}$ recorded variables, between one and
four direct causes ($|D|$) and between 8 and 11 batch conditions ($m$), and generates ${B = 200}$ batches of ${T = 512}$
steps; the size configurations raise $p$ to 40 or lower $B$ to 25, 50 or 100 (Table~\ref{tab:dataset}). The
candidates are the $p - 1$ recorded variables other than the actuator, or $p - 2$ when the controller is on
(Section~\ref{sec:coupling}). The reference configuration has no additional component and moderate shares; the other
\NCellsOther{} configurations (Table~\ref{tab:dataset}) move one share of Section~\ref{sec:shares} or one component of
Section~\ref{sec:optional} over a range, with all else at the reference, or combine several. The \emph{combined
difficulties}, used throughout Section~\ref{sec:experiments}, move five shares at once: ${s_{\mathrm{sig}} = 0.3}$,
${s_{\mathrm{conf}} = s_{\mathrm{up}} = 0.6}$, ${s_{\mathrm{desc}} = 0.5}$ and ${s_{\mathrm{int}} = 1}$, that is, a faint
signal, strong confounding and upstream links, strong downstream measurements and causes that act only jointly.
All ranges and the reference values are listed in Appendix~\ref{app:params}. Each configuration is evaluated on 10 plants. We
audit the roles of one plant per configuration by intervention: replayed with its noise held fixed, ${\mathrm{do}(X := X + \xi)}$ on a
direct cause or an ancestor changes $\Y$, on a sibling or a descendant leaves it exactly unchanged.

\begin{table}[t]
\centering\small
\caption{The released benchmark: \NCells{} configurations of the generator.
Candidates per plant: mean $\pm$ s.d.\ over all plants. Shares: each setting, with all else at the reference, and
its value measured on the generated plants (median over 10).}
\label{tab:dataset}
{\setlength{\tabcolsep}{3pt}\begin{tabular}{@{}l llllll@{}}
\toprule
configurations & \multicolumn{6}{l}{119 (the settings scored in Section~\ref{sec:experiments})} \\
plants & \multicolumn{6}{l}{1,190 (10 per configuration)} \\
batches $\times$ time steps & \multicolumn{6}{l}{25, 50, 100 or 200 $\times$ 512} \\
recorded variables & \multicolumn{6}{l}{20 or 40} \\
candidates per plant & \multicolumn{6}{L{0.70\linewidth}}{2.6 $\pm$ 0.9 direct causes, 6.8 $\pm$ 3.4 ancestors, 9.8 $\pm$ 6.0 siblings, 2.5 $\pm$ 1.1 descendants} \\
\midrule
\emph{shares: setting $\to$ measured} & \emph{reference} & \multicolumn{5}{l}{\emph{other settings}} \\
signal $s_{\mathrm{sig}}$ & 0.7\,$\to$\,0.69 & 0.05\,$\to$\,0.02 & 0.1\,$\to$\,0.07 & 0.2\,$\to$\,0.18 & 0.4\,$\to$\,0.39 & 0.9\,$\to$\,0.88 \\
upstream $s_{\mathrm{up}}$ & 0.3\,$\to$\,0.29 & 0.1\,$\to$\,0.11 & 0.6\,$\to$\,0.59 & 0.8\,$\to$\,0.79 \\
downstream $s_{\mathrm{desc}}$ & 0.25\,$\to$\,0.25 & 0.1\,$\to$\,0.11 & 0.5\,$\to$\,0.50 \\
interaction $s_{\mathrm{int}}$ & 0\,$\to$\,0.00 & 0.25\,$\to$\,0.21 & 0.5\,$\to$\,0.42 & 0.75\,$\to$\,0.65 & 1\,$\to$\,0.82 \\
confounder $s_{\mathrm{conf}}$, odds $\times g^2$ & 1 & 0.26\,$\to$\,0.26 & 3.50\,$\to$\,3.37 & 9.33\,$\to$\,8.59 \\
\bottomrule
\end{tabular}}

\end{table}

\paragraph{Baselines.} We compare eight methods from four families: the association measures lagged correlation and
Granger causality \citep{granger1969}, which do not adjust for the other candidates; the conditional-independence
tests PC \citep{spirtes2000causation} and PCMCI+ \citep{runge2019pcmci,runge2020pcmciplus}; the score-based and
functional models DYNOTEARS \citep{pamfil2020dynotears}, VARLiNGAM \citep{hyvarinen2010varlingam} and SpaceTime
\citep{mameche2025spacetime}; and DoubleML \citep{chernozhukov2018double} with LightGBM nuisance models \citep{ke2017lightgbm}, which
estimates the effect of each candidate's past on $\Y$. Hyperparameters are the same in every configuration and listed in Appendix~\ref{app:methods}. As sortability checks, we also rank the candidates by their
variance and by how well the other variables predict them \citep{reisach2021beware,reisach2023scale,lohse2025sortability}.
All methods receive the data at the same resolution and use the same look-back, since results depend on the lag and the sampling rate \citep{shojaie2022granger}. The methods receive the mean of every $f$ steps, with
$f$ set from the smallest target decorrelation time across plants in each configuration (${f = 32}$ at the reference;
Appendix~\ref{app:identifiability}), and look back twice the largest. This matches what a
process historian normally stores: averages over fixed intervals. The data are
released at full resolution.

\paragraph{Metrics.} Each method returns a selected list of candidates, possibly empty, and a ranking of all
candidates. The selected list is what a method would hand to a process engineer, and we score it by its \emph{first pick}, whether its first
entry is a direct cause (an empty list counts as wrong); its \emph{precision}, over the plants where it is not
empty; its \emph{lift}, the precision divided by that of a random list ($|D|/|\mathcal{C}|$); its
\emph{length}; and how often it is \emph{empty}. The ranking does not depend on where a method cuts its list; we score
it with \emph{R-precision}, the share of direct causes among the first $|D|$ candidates, mean average precision,
mean reciprocal rank and nDCG@3 \citep{manning2008ir}; ties are ordered at random.

\section{Experiments}
\label{sec:experiments}

\begin{figure}[t]
\centering
\includegraphics[width=0.85\linewidth]{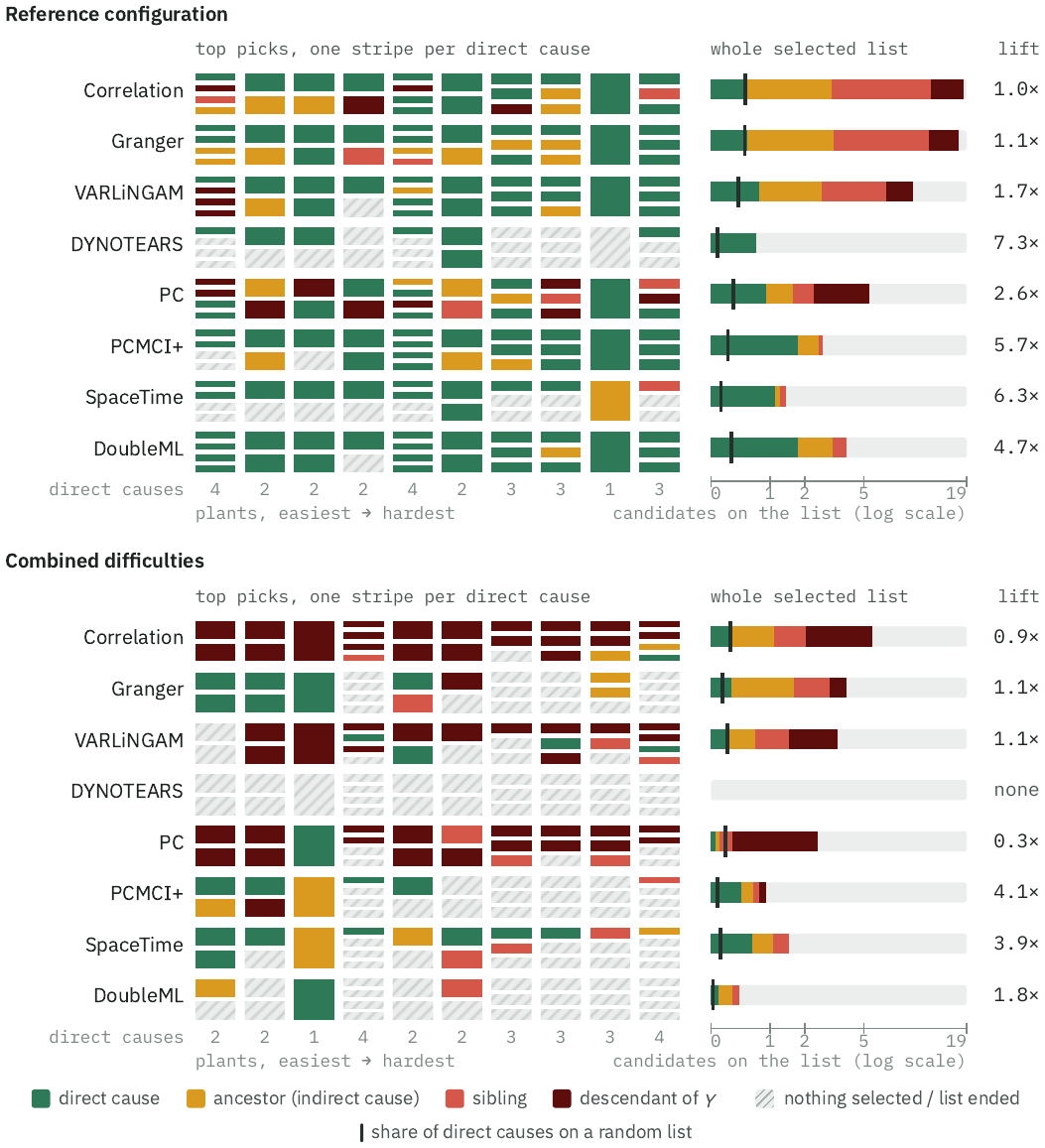}
\caption{The selected lists of the eight methods on the 10 plants of the reference configuration (top) and of the
combined difficulties (bottom). Green beyond the black tick is better than random; grey lift: fewer than three
non-empty lists.}
\label{fig:money}
\end{figure}

Figure~\ref{fig:money} shows the selected lists on the
reference configuration and under the combined difficulties, Table~\ref{tab:overall} the metrics of each method over
all \NCells{} configurations, and Appendix~\ref{app:tables} every configuration and metric. Each method selects candidates at a fixed threshold: PC and PCMCI+ at a significance level of 0.05; correlation,
Granger causality and DoubleML at a Benjamini--Hochberg adjusted $p$-value below 0.05 \citep{benjamini1995fdr};
the others by the rules in Appendix~\ref{app:methods}.

\paragraph{Reference and combined difficulties.} In the reference configuration (Figure~\ref{fig:money}, top), every method except PC ranks a direct cause first on
most plants; the methods differ in their lists. Correlation and Granger causality select
\Len{v5a}{interaction_0.0}{correlation} and \Len{v5a}{interaction_0.0}{granger} of the 19 candidates, with lifts near
one, while PCMCI+, DoubleML and SpaceTime, which adjust for the other candidates, select one to three with lifts of
\Lift{v5a}{interaction_0.0}{pcmciplus}, \Lift{v5a}{interaction_0.0}{doubleml} and
\Lift{v5a}{interaction_0.0}{spacetime_rule}.

Performance deteriorates when the difficulties are combined (Figure~\ref{fig:money}, bottom;
Table~\ref{tab:allresults}). Two difficulty measures computed from the generated data confirm that the
combined configurations are harder and place them alongside three public industrial process datasets
(Appendix~\ref{app:map}).
Correlation, VARLiNGAM and PC name a descendant of $\Y$ first on \R{v5a}{combo_interaction}{correlation}{bw},
\R{v5a}{combo_interaction}{varlingam}{bw} and \R{v5a}{combo_interaction}{pc}{bw} of 10 plants; PCMCI+, DoubleML and
DYNOTEARS select nothing on \R{v5a}{combo_interaction}{pcmciplus}{ab}, \R{v5a}{combo_interaction}{doubleml}{ab} and
\R{v5a}{combo_interaction}{dynotears}{ab}; SpaceTime names a direct cause first on
\R{v5a}{combo_interaction}{spacetime_rule}{fpn} plants with a list of \Len{v5a}{combo_interaction}{spacetime_rule}
candidates, Granger on \R{v5a}{combo_interaction}{granger}{fpn} with one of \Len{v5a}{combo_interaction}{granger}.

\paragraph{One difficulty at a time.} Because the generator sets each component independently, through the shares of Section~\ref{sec:shares} and the
additional components of Section~\ref{sec:optional}, we can simulate configurations that depart from the reference in
one component only (Table~\ref{tab:allresults}; Figure~\ref{fig:money-appx} shows two). A faint signal (${s_{\mathrm{sig}} = 0.1}$) makes the downstream measurements, which respond to all of $\Y$
including the part no recorded variable explains, more strongly associated with $\Y$ than its causes: correlation
names a descendant first on \R{v4h}{signal_0.1}{correlation}{bw} plants, while Granger and PCMCI+, which use the delay,
still name a direct cause first on most; SpaceTime, which accepts a parent only if it shortens the description length of $\Y$,
selects nothing on \R{v4h}{signal_0.1}{spacetime_rule}{ab} plants. Joint effects (${s_{\mathrm{int}} = 1}$) give each
cause an average linear effect near zero: DoubleML, whose partially linear model estimates that quantity, abstains on
\R{v5a}{interaction_1.0}{doubleml}{ab} plants, while SpaceTime, whose Gaussian-process mechanisms represent joint
effects, names a direct cause first on \R{v5a}{interaction_1.0}{spacetime_rule}{fp}. Stronger confounding pulls the
siblings ahead of the weaker causes: at ${s_{\mathrm{conf}} = 0.8}$, correlation, VARLiNGAM, DYNOTEARS and SpaceTime
name a direct cause first on \R{v4h}{confound_0.8}{correlation}{fpn}, \R{v4h}{confound_0.8}{varlingam}{fpn},
\R{v4h}{confound_0.8}{dynotears}{fpn} and \R{v4h}{confound_0.8}{spacetime_rule}{fpn} of 10 plants, while PCMCI+ and
DoubleML, which condition on the direct causes, keep \R{v4h}{confound_0.8}{pcmciplus}{fpn} and
\R{v4h}{confound_0.8}{doubleml}{fpn}. Upstream links, downstream measurements
and the controller leave PCMCI+, DoubleML and SpaceTime close to their reference results
(Table~\ref{tab:allresults}, rows $s_{\mathrm{up}}$, $s_{\mathrm{desc}}$ and $\kappa$).

The batch-level components and the imperfections generally leave the first picks of PCMCI+, DoubleML and SpaceTime
intact. Sensor error (error share 0.5) lowers the precision of the methods that condition on
the candidates, since a noisy reading does not fully block a path (PCMCI+: \R{v5a}{interaction_0.0}{pcmciplus}{prec}
$\to$ \R{v6}{meas_0.5}{pcmciplus}{prec}). Unrecorded product grades weaken the pooled association of each cause with
$\Y$ but not that of its descendants: correlation names a descendant first on \R{v6}{context_1.0}{correlation}{bw}
plants. Lots and drift move every variable together from batch to batch, which misleads correlation
(\R{v5a}{interaction_0.0}{correlation}{fp} $\to$ \R{v9}{lot_0.9}{correlation}{fp} at lot share 0.9) but not the
methods that condition on the direct causes. Recipe phases align non-causes with $\Y$ in time and add wrong
candidates to the lists of PCMCI+ (precision \R{v5a}{interaction_0.0}{pcmciplus}{prec} $\to$
\R{v8}{phase_0.9}{pcmciplus}{prec}). Variable delays and operator reactions change little at the evaluation
resolution. The number of batches matters: with 25 instead of 200, the first picks of PCMCI+ and DoubleML worsen, while
those of Granger and SpaceTime change little.

\begin{table}[t]
\centering\footnotesize
\setlength{\tabcolsep}{3.5pt}
\caption{Each method over all configurations (top) and over those with the combined difficulties (bottom).
Subscripts: half-width of the 95\% $t$-interval over the 10 plant structures, each averaged over its configurations
(precision and lift: over the structures with a non-empty list).}
\label{tab:overall}
\begin{tabular}{@{}l ccccc cccc@{}}
\toprule
 & \multicolumn{5}{c}{selected list} & \multicolumn{4}{c}{ranking} \\
\cmidrule(lr){2-6}\cmidrule(l){7-10}
Method & first pick & precision & lift & length & \shortstack{nothing\\selected} & R-precision & mAP & MRR & nDCG@3 \\
\midrule
\multicolumn{10}{@{}l}{\textit{All configurations} (1,190 plants)} \\
\quad\textit{random} & 0.13 & 0.13 & 1.0 & \textit{n/a} & \textit{n/a} & 0.13 & 0.24 & 0.31 & 0.15 \\
\quad Correlation & \cellcolor{shade!45}0.64$_{\pm.09}$ & \cellcolor{shade!6}0.15$_{\pm.03}$ & \cellcolor{shade!6}1.1$_{\pm.1}$ & 16.9$_{\pm.8}$ & \cellcolor{shadetan!6}0.00$_{\pm.00}$ & \cellcolor{shade!27}0.47$_{\pm.09}$ & \cellcolor{shade!39}0.61$_{\pm.09}$ & \cellcolor{shade!39}0.76$_{\pm.06}$ & \cellcolor{shade!31}0.57$_{\pm.11}$ \\
\quad Granger & \cellcolor{shade!60}0.79$_{\pm.12}$ & \cellcolor{shade!10}0.21$_{\pm.06}$ & \cellcolor{shade!11}1.7$_{\pm.3}$ & 13.7$_{\pm1.7}$ & \cellcolor{shadetan!11}0.10$_{\pm.04}$ & \cellcolor{shade!41}0.59$_{\pm.13}$ & \cellcolor{shade!47}0.69$_{\pm.11}$ & \cellcolor{shade!55}0.88$_{\pm.07}$ & \cellcolor{shade!42}0.67$_{\pm.12}$ \\
\quad VARLiNGAM & \cellcolor{shade!47}0.66$_{\pm.05}$ & \cellcolor{shade!16}0.30$_{\pm.04}$ & \cellcolor{shade!17}2.3$_{\pm.7}$ & 7.6$_{\pm1.6}$ & \cellcolor{shadetan!9}0.05$_{\pm.02}$ & \cellcolor{shade!34}0.53$_{\pm.08}$ & \cellcolor{shade!41}0.63$_{\pm.07}$ & \cellcolor{shade!39}0.77$_{\pm.04}$ & \cellcolor{shade!34}0.60$_{\pm.08}$ \\
\quad DYNOTEARS & \cellcolor{shade!9}0.28$_{\pm.13}$ & \cellcolor{shade!60}1.00$_{\pm.01}$ & \cellcolor{shade!60}7.3$_{\pm3.6}$ & 0.4$_{\pm.2}$ & \cellcolor{shadetan!45}0.72$_{\pm.13}$ & \cellcolor{shade!60}0.75$_{\pm.10}$ & \cellcolor{shade!60}0.82$_{\pm.08}$ & \cellcolor{shade!60}0.92$_{\pm.03}$ & \cellcolor{shade!60}0.82$_{\pm.08}$ \\
\quad PC & \cellcolor{shade!6}0.25$_{\pm.22}$ & \cellcolor{shade!16}0.30$_{\pm.03}$ & \cellcolor{shade!16}2.3$_{\pm1.0}$ & 4.5$_{\pm.8}$ & \cellcolor{shadetan!6}0.00$_{\pm.00}$ & \cellcolor{shade!6}0.29$_{\pm.03}$ & \cellcolor{shade!6}0.27$_{\pm.12}$ & \cellcolor{shade!6}0.53$_{\pm.03}$ & \cellcolor{shade!6}0.35$_{\pm.07}$ \\
\quad PCMCI+ & \cellcolor{shade!60}0.78$_{\pm.06}$ & \cellcolor{shade!40}0.69$_{\pm.06}$ & \cellcolor{shade!43}5.3$_{\pm1.5}$ & 2.4$_{\pm.4}$ & \cellcolor{shadetan!11}0.10$_{\pm.04}$ & \cellcolor{shade!58}0.74$_{\pm.06}$ & \cellcolor{shade!59}0.81$_{\pm.05}$ & \cellcolor{shade!59}0.91$_{\pm.03}$ & \cellcolor{shade!58}0.81$_{\pm.05}$ \\
\quad SpaceTime & \cellcolor{shade!44}0.63$_{\pm.14}$ & \cellcolor{shade!44}0.74$_{\pm.12}$ & \cellcolor{shade!48}5.9$_{\pm2.0}$ & 1.1$_{\pm.2}$ & \cellcolor{shadetan!14}0.16$_{\pm.08}$ & \cellcolor{shade!30}0.50$_{\pm.06}$ & \cellcolor{shade!40}0.61$_{\pm.07}$ & \cellcolor{shade!44}0.80$_{\pm.07}$ & \cellcolor{shade!36}0.61$_{\pm.08}$ \\
\quad DoubleML & \cellcolor{shade!48}0.67$_{\pm.10}$ & \cellcolor{shade!42}0.72$_{\pm.07}$ & \cellcolor{shade!44}5.5$_{\pm2.4}$ & 2.2$_{\pm.9}$ & \cellcolor{shadetan!21}0.27$_{\pm.09}$ & \cellcolor{shade!45}0.63$_{\pm.11}$ & \cellcolor{shade!49}0.71$_{\pm.10}$ & \cellcolor{shade!46}0.82$_{\pm.07}$ & \cellcolor{shade!44}0.68$_{\pm.11}$ \\
\midrule
\multicolumn{10}{@{}l}{\textit{Configurations with the combined difficulties} (230 plants)} \\
\quad\textit{random} & 0.13 & 0.13 & 1.0 & \textit{n/a} & \textit{n/a} & 0.13 & 0.25 & 0.31 & 0.14 \\
\quad Correlation & \cellcolor{shade!23}0.16$_{\pm.13}$ & \cellcolor{shade!6}0.13$_{\pm.04}$ & \cellcolor{shade!6}1.1$_{\pm.4}$ & 11.3$_{\pm1.9}$ & \cellcolor{shadetan!6}0.00$_{\pm.00}$ & \cellcolor{shade!6}0.14$_{\pm.09}$ & \cellcolor{shade!24}0.34$_{\pm.08}$ & \cellcolor{shade!10}0.40$_{\pm.09}$ & \cellcolor{shade!13}0.26$_{\pm.14}$ \\
\quad Granger & \cellcolor{shade!60}0.48$_{\pm.23}$ & \cellcolor{shade!20}0.27$_{\pm.09}$ & \cellcolor{shade!17}2.3$_{\pm.6}$ & 7.4$_{\pm2.8}$ & \cellcolor{shadetan!23}0.31$_{\pm.15}$ & \cellcolor{shade!49}0.46$_{\pm.17}$ & \cellcolor{shade!51}0.56$_{\pm.16}$ & \cellcolor{shade!60}0.72$_{\pm.15}$ & \cellcolor{shade!56}0.52$_{\pm.19}$ \\
\quad VARLiNGAM & \cellcolor{shade!19}0.12$_{\pm.07}$ & \cellcolor{shade!12}0.19$_{\pm.06}$ & \cellcolor{shade!10}1.5$_{\pm1.2}$ & 5.2$_{\pm.9}$ & \cellcolor{shadetan!11}0.10$_{\pm.06}$ & \cellcolor{shade!14}0.20$_{\pm.08}$ & \cellcolor{shade!18}0.29$_{\pm.06}$ & \cellcolor{shade!6}0.37$_{\pm.09}$ & \cellcolor{shade!8}0.23$_{\pm.10}$ \\
\quad DYNOTEARS & \cellcolor{shade!6}0.01$_{\pm.01}$ & \cellcolor{shade!60}0.67$_{\pm1.43}$ & \cellcolor{shade!60}7.6$_{\pm23.6}$ & 0.0$_{\pm.0}$ & \cellcolor{shadetan!59}0.99$_{\pm.02}$ & \cellcolor{shade!60}0.54$_{\pm.15}$ & \cellcolor{shade!60}0.63$_{\pm.15}$ & \cellcolor{shade!59}0.72$_{\pm.12}$ & \cellcolor{shade!60}0.55$_{\pm.16}$ \\
\quad PC & \cellcolor{shade!20}0.13$_{\pm.15}$ & \cellcolor{shade!11}0.18$_{\pm.07}$ & \cellcolor{shade!9}1.4$_{\pm1.4}$ & 2.7$_{\pm.4}$ & \cellcolor{shadetan!6}0.00$_{\pm.01}$ & \cellcolor{shade!11}0.18$_{\pm.06}$ & \cellcolor{shade!6}0.20$_{\pm.12}$ & \cellcolor{shade!7}0.38$_{\pm.08}$ & \cellcolor{shade!6}0.22$_{\pm.10}$ \\
\quad PCMCI+ & \cellcolor{shade!52}0.41$_{\pm.16}$ & \cellcolor{shade!45}0.52$_{\pm.13}$ & \cellcolor{shade!31}4.1$_{\pm1.6}$ & 1.2$_{\pm.3}$ & \cellcolor{shadetan!23}0.31$_{\pm.11}$ & \cellcolor{shade!48}0.45$_{\pm.10}$ & \cellcolor{shade!52}0.57$_{\pm.11}$ & \cellcolor{shade!60}0.72$_{\pm.11}$ & \cellcolor{shade!58}0.54$_{\pm.13}$ \\
\quad SpaceTime & \cellcolor{shade!52}0.41$_{\pm.17}$ & \cellcolor{shade!38}0.45$_{\pm.14}$ & \cellcolor{shade!27}3.6$_{\pm1.6}$ & 1.1$_{\pm.2}$ & \cellcolor{shadetan!13}0.13$_{\pm.10}$ & \cellcolor{shade!28}0.30$_{\pm.06}$ & \cellcolor{shade!36}0.43$_{\pm.06}$ & \cellcolor{shade!48}0.65$_{\pm.12}$ & \cellcolor{shade!36}0.40$_{\pm.10}$ \\
\quad DoubleML & \cellcolor{shade!26}0.19$_{\pm.16}$ & \cellcolor{shade!42}0.49$_{\pm.16}$ & \cellcolor{shade!34}4.4$_{\pm3.2}$ & 0.5$_{\pm.2}$ & \cellcolor{shadetan!42}0.67$_{\pm.14}$ & \cellcolor{shade!28}0.30$_{\pm.16}$ & \cellcolor{shade!32}0.41$_{\pm.13}$ & \cellcolor{shade!27}0.51$_{\pm.14}$ & \cellcolor{shade!25}0.34$_{\pm.15}$ \\
\bottomrule
\end{tabular}

\end{table}

\paragraph{Over all configurations.} The lift of the selected list separates the methods most clearly
(Table~\ref{tab:overall}): the lists of correlation and Granger
causality are barely more precise than a random one (lift \Ovr{correlation}{lift} and \Ovr{granger}{lift}), while PCMCI+, DoubleML and SpaceTime return one to three
candidates with lifts of \Ovr{pcmciplus}{lift}, \Ovr{doubleml}{lift} and \Ovr{spacetime_rule}{lift}. DYNOTEARS has
the highest precision and lift at its paper's thresholds, \Ovr{dynotears}{prec} and \Ovr{dynotears}{lift}, but on an
almost empty list: it selects nothing on \Ovr{dynotears}{none} of the plants. At thresholds ten times lower it
selects on most plants and is about as good as PCMCI+ (Appendix~\ref{app:threshold}).
Under the combined difficulties, correlation's R-precision falls to \OvrC{correlation}{rp}, about
that of a random ranking. Granger, SpaceTime and PCMCI+ name a direct cause first on \OvrC{granger}{first},
\OvrC{spacetime_rule}{first} and \OvrC{pcmciplus}{first} of the plants and select nothing on \OvrC{granger}{none},
\OvrC{spacetime_rule}{none} and \OvrC{pcmciplus}{none}; DoubleML selects nothing on \OvrC{doubleml}{none} and names a
direct cause first on only \OvrC{doubleml}{first}. Sorting the candidates by variance or
explained variance, a known artefact of simulated data
\citep{reisach2021beware,reisach2023scale,lohse2025sortability}, picks a direct cause first on \SortVarFp{} and
\SortRtwoFp{} of the plants, against \SortRandFp{} at random.

\paragraph{Errors by method.} Because the role of every candidate is known, each wrong selection can be broken down by role
(Table~\ref{tab:errors}, Appendix~\ref{app:errors}). Correlation and Granger causality select every role in about its random
proportion but differ in their first pick: correlation ignores time order and ranks a descendant of $\Y$ first on
\ErrT{correlation}{back} of the plants, Granger on \ErrT{granger}{back}, and most of Granger's wrong first picks are
ancestors one step upstream of a cause. PC, without time order, selects descendants
(\ErrT{pc}{enr_downstream_pct} of its wrong selections, against \ErrT{pool}{downstream} at random). PCMCI+, DoubleML
and SpaceTime make fewer than one wrong selection per plant; their characteristic failure is to select nothing when
difficulties combine, on \ErrT{pcmciplus}{none}, \ErrT{doubleml}{none} and \ErrT{spacetime_rule}{none} of the plants.

\section{Conclusion}
\label{sec:discussion}
\label{sec:conclusion}

We presented a synthetic benchmark for target-centric causal discovery in industrial processes. Its generator is a
two-level structural causal model inspired by batch production. Each component is switched on and set
independently, and the ground truth gives candidate roles and separation or time-order labels. Eight methods, evaluated
on \NCells{} configurations, degrade as the difficulties combine. Configurations can change one component at a time, so that a change in a method's result can be attributed to one
mechanism.

\paragraph{Using the benchmark.} First, researchers can test a new method on each
difficulty separately and on their combinations: each configuration changes named components of the reference, and
every wrong selection is labelled with the candidate's role. Second, engineers can
generate plants that resemble their own in size and difficulties and test a method on them before applying it to the
plant's data. Third, the
generator can supply training data. Foundation models for tabular prediction and causal effect estimation are
pretrained on data drawn from structural causal models \citep{hollmann2025tabpfn,qu2026tabiclv2,robertson2025dopfn};
a generator of process data could serve the same purpose for industrial processes.

\paragraph{Limitations.} The benchmark is synthetic: the generator reproduces the causal structure of a batch
process (Section~\ref{sec:scm}), not its physics. The profiles come
from one family of motifs and the exogenous noise is Gaussian. Each configuration is evaluated on 10 plants only, and
the methods run at one resolution and look-back. The benchmark gives no finite-sample recovery bound, so a failure of
every method cannot be distinguished from an unrecoverable signal.

\paragraph{Use of generative AI.} Generative AI tools were used for language editing; the authors take full
responsibility for the content.

\bibliography{references}

\clearpage
\appendix
\section{The generator in detail}
\label{app:generator}
\label{app:optional}
\label{app:params}

This appendix details the generator of Section~\ref{sec:generator}: one entry per component with its mechanism and
equation, then the order in which a batch is generated, the parameter ranges and the interventional audit.

\paragraph{Core components.}
\begin{itemize}
  \item \emph{Batch conditions.} The $m$ latent conditions $U_1,\dots,U_m$ are linked in a random tree
  (Table~\ref{tab:params}). A condition without parents is standard Gaussian; any other is a weighted sum of its
  parents plus Gaussian noise. The conditions enter the profile parameters with random coefficients that no share
  sets, so in Equation~\ref{eq:share} they belong to the remainder $r$.
  \item \emph{Confounders.} After the direct causes are drawn, every other recorded variable, the actuator included,
  is assigned to one of them: the $p - |D|$ variables are sorted by name and split into consecutive blocks of
  $\lceil (p-|D|)/|D| \rceil$, one per direct cause, the last block taking the remainder (Table~\ref{tab:params}). Each direct cause $X_k$ then receives a latent confounder $Z_k$, a standard
  Gaussian drawn anew in each batch, which enters the profile parameters of $X_k$ and of every variable in its block
  with the confounder share $s_{\mathrm{conf}}$. A variable in the block of $X_k$ therefore shares a latent cause with
  $X_k$. It is a sibling of $\Y$ unless it is also an ancestor or a descendant of $\Y$, in which case it keeps that
  role. Its within-batch component $Z_k(t)$, white noise smoothed with a Gaussian filter, is added to the
  trajectories of $X_k$ and of its block.
  \item \emph{Profile parameters.} Each parameter $\theta$ is set from its parents $\pa(\theta)$ by
  \begin{equation}
  \label{eq:param}
  \theta = \ell_\theta + (h_\theta - \ell_\theta)\,\varsigma\Big(\textstyle\frac{1}{|\pa(\theta)|}\sum_{q \in \pa(\theta)} q + \sigma_\theta\,\varepsilon_\theta\Big),
  \end{equation}
  where $\varsigma$ is the logistic function, $[\ell_\theta, h_\theta]$ the admissible range of $\theta$
  (Table~\ref{tab:params}) and $\varepsilon_\theta$ standard Gaussian noise. The parents are batch conditions,
  confounders, phase timings and parameters of other recorded variables (the batch-level links); the weights are
  equal except on the routes that carry a share.
  \item \emph{Profiles.} A recorded variable $X_i$ has $K_i \in \{2,3,4\}$ motifs, each with a direction
  $d_k \in \{-1,+1\}$ (up or down); their shapes $\mu_k$, directions and order are drawn once per plant and then
  fixed. In each batch,
  \begin{equation}
  \label{eq:trajectory}
  X_i(t) = \lambda_i + \sum_{k=1}^{K_i} \phi_k\, d_k\, A_k\, \psi_{\mu_k}(t; o_k, \delta_k) + \eta_i(t) + \nu_i(t),
  \end{equation}
  where the level $\lambda_i$, the sizes $A_k$, the onsets $o_k$ (each within the slot of its motif) and the
  durations $\delta_k$ are the profile parameters of the batch, $\psi_{\mu_k}$ is the shape of motif $k$,
  $\phi_k \in \{0,1\}$ indicates whether the motif occurs in this batch, $\eta_i$ is a smooth random texture (a
  Gaussian process) and $\nu_i$ is white noise. The four shapes are two changes that persist (a smooth step and a
  linear ramp) and two that return (a spike and a plateau with short edges), each with onset $o$ and duration
  $\delta$:
  \begin{align*}
  \text{step (persists):}\quad & \psi(t; o, \delta) = \big(1 + e^{-(t - o - \delta/2)/(0.15\,\delta)}\big)^{-1}, \\
  \text{ramp (persists):}\quad & \psi(t; o, \delta) = \min\!\big(1, \max(0, (t - o)/\delta)\big), \\
  \text{spike (returns):}\quad & \psi(t; o, \delta) = \exp\!\big(-\tfrac12 \big((t - o - \delta/2)/(\delta/4)\big)^2\big), \\
  \text{plateau (returns):}\quad & \psi(t; o, \delta) = \min\!\big(h(t - o),\, h(o + \delta - t)\big),
  \end{align*}
  with $h(u) = \min\!\big(1, \max(0, u/(0.15\,\delta))\big)$. In a batch plant, a heat-up looks like a step, a dosing
  phase like a ramp, an exotherm (the temperature spike from the heat released by the reaction) like a spike and a
  holding phase like a plateau; the generator draws the shapes at random. In each
  batch a motif occurs with a fixed probability, and its onset and size are jittered.
  \item \emph{Time-step links.} A variable $X_j$ with upstream variables receives the sum of their delayed responses,
  mixed with its trajectory before the links, $X_j^{0}$, as in Equation~\ref{eq:share}:
  \begin{equation*}
  X_j(t) = \sqrt{s_{\mathrm{up}}}\,\tilde a_j(t) + \sqrt{1-s_{\mathrm{up}}}\,\tilde X_j^{0}(t), \qquad
  a_j(t) = \sum_{i} \beta_{ij}\, \mathrm{F}_{\tau_{ij},\vartheta_{ij}}[X_i](t),
  \end{equation*}
  with $\mathrm{F}$ the first-order-plus-dead-time response of Section~\ref{sec:coupling}. Variables are coupled in
  causal order, so each receives its upstream variables already coupled; the standardising scales are fixed under
  intervention. The gain $\beta_{ij}$, time constant $\tau_{ij}$ and dead time $\vartheta_{ij}$ of each link are drawn
  once per plant (Table~\ref{tab:params}).
  \item \emph{Controller.} The controller holds $X_c$ at the setpoint $\rho(t)$ against the delayed disturbance $X_d$
  by moving the actuator $X_a$,
  \begin{equation}
  \label{eq:loop}
  \begin{aligned}
  X_c(t) &= \rho(t) + \mathrm{F}_{\tau_d,\vartheta_d}[X_d](t) + X_a(t-1), \\
  u(t) &= u(t-1) + \kappa\big(\rho(t) - X_c(t)\big), \qquad X_a(t) = X_a^{0}(t) + u(t),
  \end{aligned}
  \end{equation}
  with $u(t)$ the accumulated correction and $X_a^{0}$ the actuator's own profile.
  \item \emph{Target.} The interaction terms are products of pairs of direct causes, chosen so that every direct
  cause is in at least one pair; each product has its own gain. The unexplained part $\varepsilon$ of $\Y$
  (Equation~\ref{eq:y}) stands for causes of $\Y$ that are not recorded. Like the recorded causes, these change
  slowly, so $\varepsilon$ is drawn as a slowly varying random process: an AR(1) whose autocorrelation falls to $1/e$
  at the decorrelation time of $S$. If $\varepsilon$ were white noise, the block means the methods receive would
  average it away, and the direct causes would explain more of $\Y$ than $s_{\mathrm{sig}}$ sets.
  \item \emph{Downstream measurements.} Each receives the delayed response to $\Y$, with a dead time set from the
  decorrelation time of $\Y$, mixed with its own profile by the downstream share $s_{\mathrm{desc}}$.
\end{itemize}

\paragraph{Additional components.}
\begin{itemize}
  \item \emph{Recipe phases.} The batch is divided into four equal phase windows. Each window $q$ has a latent timing
  $\Phi_q$, drawn once per batch, that states how early or late the batch runs through that phase; it is a parent, in
  Equation~\ref{eq:param}, of the onset of every motif whose slot lies in the window, so the motifs of all variables
  in that window shift together. The weight of $\Phi_q$ is set so that it carries the share $s_{\mathrm{phase}}$ of
  each such onset's variance: before the data are drawn, the generator simulates 3,000 batches of the profile
  parameters, measures the variance that the onset's other parents contribute and solves for the weight. The equation
  is linear, so the share holds exactly. The weights of the confounders are set in the same way.
  \item \emph{Lots and drift.} Both are set on each variable's \emph{batch-to-batch variation}, the variance over
  batches of its batch mean $\bar X_{i,b}$; this includes everything that changes between batches (the profile
  parameters, whether a motif occurs, its timing and size). The lot process $L_b$ is
  an AR(1) over the batch index and the drift $R_b$ a linear trend in it; the lot models consecutive batches
  sharing a raw-material lot (persistent but random, with a random sign $\zeta_i$ per variable), the drift a slow
  change over the campaign such as fouling or catalyst ageing (the same direction for every variable); each variable's trajectory is shifted by a
  constant per batch such that, in standardised units,
  \begin{equation}
  \label{eq:campaign}
  \bar X_{i,b} \;\to\; \sqrt{1 - s_{\mathrm{lot}} - s_{\mathrm{drift}}}\;\tilde{\bar X}_{i,b}
    + \sqrt{s_{\mathrm{lot}}}\, \zeta_i\, \tilde L_b + \sqrt{s_{\mathrm{drift}}}\, \tilde R_b ,
  \end{equation}
  with a sign $\zeta_i$ per variable, as in Equation~\ref{eq:share}. The shape of the trajectory within the batch is
  unchanged, and the shift is applied before the time-step links, the controller and the target, so lots and drift
  reach $\Y$ only through its direct causes.
  \item \emph{Sensor error.} The methods receive readings, not true values: each variable is recorded as
  $X_i^{\mathrm{rec}} = \sqrt{1-s_{\mathrm{err}}}\,\tilde X_i + \sqrt{s_{\mathrm{err}}}\,\tilde e_i$, so the error
  carries the share $s_{\mathrm{err}}$ of the reading's variance. The error $e_i$ varies slowly, like a drifting
  sensor: an AR(1) with the decorrelation time of $X_i$, so that the block means do not average it away. The plant
  itself runs on the true values; the links, the controller and the target never see the error.
  \item \emph{Product grades.} The grade $\gamma \in \{-1,+1\}$ of each batch is drawn with equal probability and is
  not recorded. It multiplies the weight of each direct cause by $(1 + c\,\zeta_k \gamma)/\sqrt{1+c^2}$, where
  $c \in [0,1]$ is the grade contrast (Section~\ref{sec:optional}) and the sign $\zeta_k \in \{-1,+1\}$ of each cause
  selects the grade in which that cause acts more strongly. At $c = 0$ the two grades are identical; at $c = 1$ each
  cause acts in one grade only. The squared multiplier averages to one over the two grades, so the average strength
  of each cause does not depend on $c$.
  \item \emph{Variable delays.} In each batch, the dead time $\vartheta_k$ of each direct cause is increased by a
  hold-up drawn uniformly between zero and $j$ times the decorrelation time of the cause, so the response of $\Y$ to
  the cause falls at different lags from batch to batch.
  \item \emph{Cumulative effects.} Some causes act through their accumulated exposure, e.g.\ the heat a product has
  received over the last hour, and not only through their current value. The response of $\Y$ to a cause is
  therefore a mix of two parts, both standardised and combined as in Equation~\ref{eq:share} with weight
  $k_{\mathrm{avg}}$: the cause's current delayed response, and its running mean over the last two decorrelation
  times of the cause. At $k_{\mathrm{avg}} = 0$ only the current value acts; at $k_{\mathrm{avg}} = 1$ only the
  accumulated one. The two parts are correlated, so $k_{\mathrm{avg}}$ sets their relative weight but is not the
  share of the response's variance that each carries (Section~\ref{sec:shares}).
  \item \emph{Operator reactions.} An operator watches $\Y$ and corrects it through one direct cause, chosen among
  those that act on $\Y$ monotonically (linear, saturating or thresholded), so that countering a deviation has a
  direction. The operator sees $\Y$ through a first-order lag with time constant $\tau_{\Y}$, delayed by $\tau_{\Y}$,
  the decorrelation time of $\Y$, and moves the reference value of the cause against it: the cause becomes
  $\sqrt{1-s_{\mathrm{op}}}$ times its own trajectory plus $\sqrt{s_{\mathrm{op}}}$ times this reaction, both
  standardised as in Equation~\ref{eq:share}. Over the variables, $\Y \to X \to \Y$ is a cycle; over time it is not,
  since the cause at time $t$ depends only on $\Y$ up to $t - \tau_{\Y}$. The generator therefore computes the batch
  forward in steps of $\tau_{\Y}$, each fixed by the previous ones; Appendix~\ref{app:identifiability} treats the
  cycle in the separation labels.
\end{itemize}

\paragraph{Order of generation.} Algorithm~\ref{alg:episode} gives the order in which one batch is generated.

\begin{algorithm2e}[h]
\caption{Generation of one batch of a plant.}
\label{alg:episode}
\DontPrintSemicolon
\small
Draw the batch conditions $U_j$, the confounders $Z_k$ and the phase timings $\Phi_q$\;
\For{each profile parameter $\theta$, in causal order}{
  $\theta \leftarrow$ Equation~\ref{eq:param}, from its parents among the $U_j$, $Z_k$, $\Phi_q$ and the
  parameters of other variables\;
}
\For{each recorded variable $X_i$}{
  $X_i(t) \leftarrow$ its profile with timing and sizes from $\theta(X_i)$, plus texture (Equation~\ref{eq:trajectory})\;
}
Add the within-batch part $Z_k(t)$ of each confounder to its direct cause and to the variables of its block\;
Shift each variable by its lot and drift contribution for this batch (Equation~\ref{eq:campaign})\;
\For{each recorded variable with upstream variables, in causal order}{
  mix in the delayed responses of its upstream variables (time-step links)\;
}
Close the control loop (Equation~\ref{eq:loop})\;
$\Y(t) \leftarrow$ response to the direct causes, with product grades, variable delays and cumulative effects if
set, plus the unexplained part (Equations~\ref{eq:target} and~\ref{eq:y})\;
\If{operator reactions are set}{
  move the reference value of one direct cause against the past of $\Y$ and recompute $\Y$, forward in steps of
  $\tau_{\Y}$\;
}
Add the delayed response to $\Y$ to each downstream measurement\;
\If{sensor error is set}{
  record each variable as a mix of its true value and its sensor error\;
}
\end{algorithm2e}

\paragraph{Parameters and seeds.} Table~\ref{tab:params} lists the parameter ranges used in all experiments. Values
are drawn once per plant unless marked per batch; time is in steps. Each plant has two seeds: a structure seed, which
draws the plant, and a data seed, which draws only its batches, so datasets drawn with different data seeds share the
plant's structure.

\begin{table}[h]
\centering\small
\caption{Parameter ranges of the generator in all experiments.}
\label{tab:params}
\begin{tabular}{@{}p{0.34\linewidth} p{0.58\linewidth}@{}}
\toprule
Quantity & Value \\
\midrule
\multicolumn{2}{@{}l}{\emph{Batch level}} \\
batch conditions $U_j$ & $m =$ 8--11 (log-uniform); random directed tree, built by adding the conditions one by one, each linked to a random earlier one or, with a small probability, to a neighbour of it, which creates a few hubs (growing network with redirection, \citealp{krapivsky2001organization}); conditions without parents $\mathcal{N}(0,1)$; weights uniform in $[-1,1]$, noise s.d.\ $\sigma_U = 0.2$ \\
confounders $Z_k$ & one per direct cause, $\mathcal{N}(0,1)$; the $p - |D|$ recorded variables that are not direct causes, the actuator included, are assigned to the causes in blocks (19, 9, 5--6 or 4 per cause at $p = 20$ for $|D| = 1$ to 4); $Z_k$ enters each profile parameter of a variable in its block with probability 0.7 \\
profile parameters (Eq.~\ref{eq:param}) & equal weights $1/|\pa(\theta)|$ except on routes with a share; noise s.d.\ $\sigma_\theta = 0.2$; onset 0--0.5 and duration 0.2--0.5 of the slot; size 0.5--2 (times a base size 1.5); level $-0.5$--$0.5$ \\
batch-level links & expected in-degree 1; 2 upstream variables \\
lots and drift (Eq.~\ref{eq:campaign}) & lot process: AR(1) over batches, persistence 0.9; drift: linear in the batch index \\
\multicolumn{2}{@{}l}{\emph{Profiles}} \\
motifs per variable $K_i$ & 2--4, shape and direction uniform \\
per-batch variation & motif occurs with probability 0.7; onset jitter up to $\pm 16$ steps; size jitter 15\% \\
texture $\eta_i$ & amplitude 0.2--1.0; length scale 0.005--0.05 of $T$ \\
white noise $\nu_i$ & s.d.\ 0.1 \\
\multicolumn{2}{@{}l}{\emph{Time-step level}} \\
time-step links & gain $\pm$0.3--0.8; $\tau$ 3--20; $\vartheta$ 0--5 \\
within-batch confounder $Z_k(t)$ & white noise smoothed with a Gaussian filter of width 60 steps, unit variance in
each batch; strength per variable uniform in $[0.3, 1.0]$, times $g$ (Appendix~\ref{app:realised}) \\
controller (Eq.~\ref{eq:loop}) & one controller on a direct cause; disturbance $\tau_d = 10$, $\vartheta_d = 3$; $\kappa = 0$ unless varied; with the controller on, the disturbance of the controlled cause is not a candidate, leaving 18 \\
\multicolumn{2}{@{}l}{\emph{Target}} \\
direct causes $|D|$ & 1--4 (4--6 in one configuration) \\
links to the target (Eq.~\ref{eq:target}) & $|w_k|$ 0.6--1.2, random sign; $\tau_k$ 1--30; $\vartheta_k$ 0--15; $g_k$ uniform over the four response functions \\
interactions & pairs of direct causes that together cover $D$ \\
descendants of $\Y$ & 2--3; dead time 1--2 decorrelation times of $\Y$ \\
\multicolumn{2}{@{}l}{\emph{Reference configuration}} \\
shares and controller & $s_{\mathrm{sig}} = 0.7$, $s_{\mathrm{conf}} = s_{\mathrm{up}} = 0.3$, $s_{\mathrm{desc}} = 0.25$, $s_{\mathrm{int}} = 0$, $\kappa = 0$; no additional component \\
\bottomrule
\end{tabular}
\end{table}

\paragraph{Realised shares.}\label{app:realised} Table~\ref{tab:realised} compares every share with its value measured
on the generated plants, at the resolution and look-back the methods use, as an out-of-sample $R^2$ (five folds split
by batch; ridge regression on the standardised lags $0,\dots,L$), on the 10 plants of each setting with all else at
the reference. The signal share is the $R^2$ of $\Y$ from its direct causes, with the squares of their lags and the
products of pairs, so that nonlinear and joint effects count; the upstream share is the $R^2$ of a variable from the
variables upstream of it; the downstream share is the $R^2$ of a downstream measurement from $\Y$; the interaction
share is the part of the causes' $R^2$ that only the products carry. The confounder share acts in two stages. Its
share of the profile parameters is set exactly by calibration, but the parameters are not recorded. Its within-batch
component $Z_k(t)$ is added to the trajectory of the cause and of each sibling with a strength drawn per variable
(Table~\ref{tab:params}), multiplied by $g = \sqrt{s/(1-s)}\,/\sqrt{s_0/(1-s_0)}$ with $s_0 = 0.3$, so that the
within-batch odds $q/(1-q)$ of each variable change by the factor $g^2$ from the reference, where $q$, the share of the
variable's within-batch variance due to $Z_k(t)$, is measured as $1 - \mathrm{corr}^2$ between the variable with and without $Z_k(t)$, all else fixed (\ConfPlants{} plants per
setting). Three measured values fall below their settings, each for a known reason. A faint signal
($s_{\mathrm{sig}} \le 0.1$) is estimated low, as an out-of-sample $R^2$ is when the signal is weak. At high
interaction shares, the squares of the causes in the regression absorb part of the joint effect, so the share that
only the products carry reads low. At $s_{\mathrm{conf}} = 0.8$, part of the within-batch component is passed on to
other variables by the time-step links or countered by the controller, so less of it remains in the variable than
was added. Siblings are also associated with their cause through the shared batch conditions: with both stages
of the confounder removed, the $R^2$ a sibling shares with its cause is \SibFloor{}, against \SibRef{} at the
reference (median over \SibPlants{} plants).

\begin{table}[h]
\centering\small
\caption{Configured and realised shares: each setting, with all else at the reference (in bold), against its value
measured on the generated plants (median and interquartile range). For the confounder share, the measured quantity is
the factor by which the within-batch odds of a cause or sibling change from the reference, designed to be $g^2$.}
\label{tab:realised}
\begin{tabular}{@{}l l l l@{}}
\toprule
Share & Setting & Expected & Measured, median [IQR] \\
\midrule
$s_{\mathrm{sig}}$ & 0.05 & 0.05 & 0.02 [0.01, 0.04] \\
 & 0.1 & 0.1 & 0.07 [0.06, 0.10] \\
 & 0.2 & 0.2 & 0.18 [0.16, 0.21] \\
 & 0.4 & 0.4 & 0.39 [0.36, 0.42] \\
 & \textbf{0.7} & 0.7 & 0.69 [0.68, 0.71] \\
 & 0.9 & 0.9 & 0.88 [0.87, 0.89] \\
\addlinespace[2pt]
$s_{\mathrm{up}}$ & 0.1 & 0.1 & 0.11 [0.10, 0.14] \\
 & \textbf{0.3} & 0.3 & 0.29 [0.26, 0.34] \\
 & 0.6 & 0.6 & 0.59 [0.57, 0.61] \\
 & 0.8 & 0.8 & 0.79 [0.78, 0.80] \\
\addlinespace[2pt]
$s_{\mathrm{desc}}$ & 0.1 & 0.1 & 0.11 [0.09, 0.15] \\
 & \textbf{0.25} & 0.25 & 0.25 [0.24, 0.32] \\
 & 0.5 & 0.5 & 0.50 [0.48, 0.56] \\
\addlinespace[2pt]
$s_{\mathrm{int}}$ & \textbf{0} & 0 & 0.00 [-0.01, 0.00] \\
 & 0.25 & 0.25 & 0.21 [0.14, 0.24] \\
 & 0.5 & 0.5 & 0.42 [0.31, 0.47] \\
 & 0.75 & 0.75 & 0.65 [0.50, 0.71] \\
 & 1 & 1 & 0.82 [0.66, 0.95] \\
\addlinespace[2pt]
$s_{\mathrm{conf}}$ & 0.1, $g = 0.51$ & $g^2 = 0.26$ & 0.26 [0.26, 0.27] \\
 & \textbf{0.3}, $g = 1.00$ & $g^2 = 1.00$ & 1 (by definition) \\
 & 0.6, $g = 1.87$ & $g^2 = 3.50$ & 3.37 [3.21, 3.48] \\
 & 0.8, $g = 3.06$ & $g^2 = 9.33$ & 8.59 [7.69, 9.21] \\
\bottomrule
\end{tabular}

\end{table}

\paragraph{Interventional audit.} The roles are read on the generating graph; the audit checks them against the
generator's behaviour on one plant of every configuration (\NCells{} plants, 10\%). (i) The plant is rebuilt from its configuration and seeds and replayed;
the replayed $\Y$ must equal the released one. (ii) For every recorded variable, one intervention at a time with all
exogenous noise held fixed: its trajectory is shifted, $\mathrm{do}(X_i := X_i + \xi)$ with $\xi$
white noise of the variable's standard deviation, and each of its profile parameters is shifted by one standard
deviation and propagated through the SCM. The effect is the root-mean-square change of $\Y$ in units of its standard
deviation. (iii) A direct cause or an ancestor must have a non-zero effect; a sibling or
a descendant must have an effect of exactly zero (below $10^{-12}$); and every profile parameter must change the
trajectory of its own variable, since a parameter without effect would be a false edge.

\section{Ground truth and separation labels}
\label{app:fullgraph}
\label{sec:extract}
\label{app:identifiability}

\paragraph{The full graph.} The generator builds and exports the full causal graph of every plant at three levels:
the fine-grained structural causal model, with every latent batch condition, confounder, phase timing, profile
parameter and trajectory as a node (about \LabRefNodes{} nodes for 20 recorded variables); its latent projection onto the recorded
variables, with a directed edge where one variable acts on another directly or through latent nodes only and a
bidirected edge where two share a latent cause; and the links between recorded variables, split into batch-level links and
time-step links with their gains, time constants and dead times.

\paragraph{Latent variables.} The batch conditions and confounders are left out of the data by choice: a plant
rarely records ambient conditions, utilities, raw-material properties or the crew on duty, although they act on every
batch. Because the batch conditions, linked by their tree, act on the profiles of every recorded variable, almost every
two recorded variables share an unrecorded cause: of the \LabRefPairs{} pairs formed by the \LabRefVars{} recorded
variables of a reference plant, at least \LabRefConfPairs{} do in every plant. This is the trade-off of a realistic plant: the full graph is hard to recover from such data, while every
wrong candidate remains separable from the direct causes of the target without sensor error.

\begin{figure}[h]
\centering
\includegraphics[width=\linewidth]{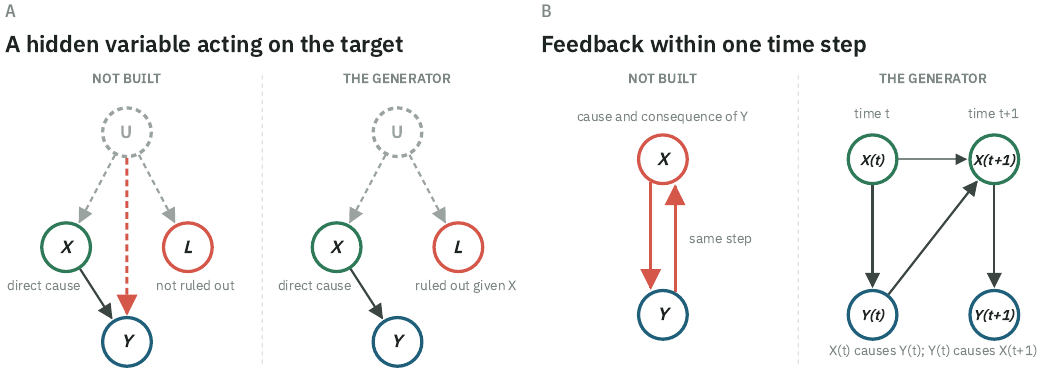}
\caption{Two structures that break separability (left in each panel) and the corresponding structures of the
generator (right). (a) An unrecorded variable $U$
acting on the target directly: conditioning on the direct cause $X$ does not separate the sibling $L$ from $\Y$.
In the generator, unrecorded variables reach $\Y$ only through its direct causes. (b) $X$
acting on $\Y$ and $\Y$ back on $X$ within one time step, as a controller or an operator reacting at once would: $X$
is both a cause and a consequence of $\Y$ and has no role. In the generator, every feedback loop acts with a delay of
at least one step, so that, unrolled in time, $X(t)$ is a cause of $\Y(t)$ and $\Y(t)$ a cause of $X(t+1)$.}
\label{fig:hidden}
\end{figure}

\paragraph{The benchmark is target-centric.} Separability depends on the structure of the plant.
Figure~\ref{fig:hidden} shows two examples of structures that break it: an unrecorded variable acting on $\Y$
directly, which leaves a path from a sibling to $\Y$ open given the direct causes, and feedback within one time step,
which makes a variable at once a cause and a consequence of $\Y$. The generator is built so that every wrong
candidate is separable from $\Y$, given the direct causes or by time order, and the label of each candidate, computed
on the graph of its plant, verifies this. In the reference
configuration all
\LabRefNonCause{} siblings over \LabRefInst{} plants are confounded with the target, yet all \LabRefWrong{} wrong
candidates have one of these labels: \LabRefIdCond{} by conditioning and \LabRefIdTime{} by time order. Sensor
error can prevent separation given the recorded causes. The difficulty is set around the target as
well: the strength of its causes, the confounders they share with siblings, the controller on one of them and the
measurements downstream of it. The full graph is released with every
dataset and can be used to evaluate full-graph causal discovery, with a metric that accounts for hidden common causes.

\paragraph{Resolution and look-back.} Both are set from the data alone, through the memory of $\Y$: for each plant,
$\tau_{\Y}$ is the lag at which the autocorrelation of $\Y$ within a batch, normalised by its lag-1 value, falls below
$1/e$. The methods receive the mean of every $f$ steps, with $f$ the largest power of two not above the smallest
$\tau_{\Y}$ in the configuration, and we verify that every descendant of $\Y$ responds to it with a delay of at least
$f$ steps, so that after averaging its response still follows $\Y$ by at least one averaged point. The methods look
back $L = 2\tau_{\Y}$ of the slowest plant, and every cause acts within that window. Both hold in every
configuration. The rule for $f$ gives $f = 32$ in all configurations except two with 50 batches, where the estimate of
the smallest $\tau_{\Y}$ falls below 32 and $f = 16$; the look-back is 3 to 8 blocks, 4 at the reference. Averaging
leaves \SepAll{} of the wrong candidates not separable from the direct causes by a linear test, against \SepFine{} at
a resolution of 4 steps.

\paragraph{Separation labels.} For every wrong candidate, we record whether a method could rule it out in
principle, computed on the true graph. It is labelled \emph{by conditioning} when it is $d$-separated from $\Y$ given
the direct causes $D$ \citep{spirtes2000causation}: once the direct causes are known, the candidate carries no further
information about $\Y$. It is labelled \emph{by time order} when it is a descendant that responds to $\Y$ at least one
time step later at the evaluation resolution (Section~\ref{sec:protocol}), and \emph{neither criterion} otherwise.
Outside the configurations with sensor error, every label is by conditioning or by time order. The test runs on the
summary graph, with one node per variable grouping its trajectory and profile parameters. A controller closes a loop
in this graph, and $d$-separation can then give wrong answers; we use $\sigma$-separation, its counterpart for cyclic
causal models \citep{forre2017markov,forre2018constraint}, computed as $d$-separation on an acyclic version of the
graph in which the variables of each loop share common causes \citep{mooij2020cycles}. An operator who
reacts to $\Y$ closes a loop only over time, since the reaction follows past values of $\Y$; the separation test
therefore conditions on the direct causes and on the past of $\Y$, as every lag-based method does, and the loop leaves
the labels unchanged. Under sensor error, a
method conditions on the readings, not on the true values, and conditioning on a noisy reading does not remove all
of the information the true value carries \citep{scheines2016measurement}. We therefore also compute the labels on the
graph in which each true value is unrecorded and only its reading is recorded; separation given the true causes
need not hold given their readings. Both versions are recorded.

\paragraph{Separability at the evaluation resolution.} The methods see block means of $f$ steps
(Section~\ref{sec:protocol}). Averaging hides the movement of a direct cause within a block, so a wrong candidate that
drives or follows it closely can still carry information about $\Y$ given the averaged causes
\citep{gong2017aggregated}. We therefore test every wrong candidate on every plant on the data the methods see: does it
improve the prediction of $\Y$ beyond the direct causes? We fit two linear regressions of $\Y(t)$ on the block means,
pooled over the batches of the plant: the first on the past of $\Y$ (lags 1 to $L$) and on the direct causes (lags 0
to $L$), with $L$ the common look-back of Section~\ref{sec:protocol}; the second adds the candidate at the same lags.
An $F$-test compares the two fits. A candidate is \emph{separable at the evaluation resolution} if it adds no
significant improvement ($p \ge 0.01$), so that a linear conditional-independence test, like those of the baselines,
would rule it out given the direct causes. Each candidate that is
not separable is tested again at a resolution of 4 steps, close to the full resolution, with the same look-back: if it
is separable there, averaging is the reason. The remaining candidates fail the linear test for other reasons, such as
a nonlinear response, sensor error, or chance. Table~\ref{tab:sep-roles} gives the shares by role over the
\SepN{} wrong candidates, and Table~\ref{tab:sep-methods} the share of each method's wrong selections that are not
separable, with the list rescored without them. The label is released with every plant.

\begin{table}[h]
\centering\small
\caption{Wrong candidates not separable from the direct causes at the evaluation resolution, by role, over all plants.
The same test at a resolution of 4 steps, close to the full resolution, splits them into two groups: \emph{by
averaging}, candidates separable at 4 steps ($p \ge 0.05$), and \emph{also at 4 steps}, candidates not separable
there either.}
\label{tab:sep-roles}
\begin{tabular}{@{}l r ccc@{}}
\toprule
Wrong candidates & number & not separable & by averaging & also at 4 steps \\
\midrule
descendants & 3,117 & 14\% & 12\% & 2\% \\
disturbance of the controlled cause & 1,150 & 23\% & 18\% & 5\% \\
other ancestors & 7,190 & 9\% & 7\% & 2\% \\
siblings & 12,010 & 7\% & 4\% & 3\% \\
\midrule
all & 23,467 & 9\% & 7\% & 2\% \\
\bottomrule
\end{tabular}

\end{table}

\begin{table}[h]
\centering\small
\setlength{\tabcolsep}{3pt}
\caption{Wrong selections of each method that are not separable at the evaluation resolution. Left: their share of
the method's wrong selections, split as in Table~\ref{tab:sep-roles} into \emph{by averaging} and \emph{also at 4
steps}, and their share of all its selections (\emph{share of list}). Right: first pick, precision and lift as in
Table~\ref{tab:overall}, then recomputed with the non-separable candidates removed from the candidates of each plant.}
\label{tab:sep-methods}
\begin{tabular}{@{}l cccc ccc@{}}
\toprule
 & \multicolumn{4}{c}{wrong selections not separable} & \multicolumn{3}{c}{as scored $\to$ without them} \\
\cmidrule(lr){2-5}\cmidrule(l){6-8}
Method & all & by averaging & also at 4 steps & share of list & first pick & precision & lift \\
\midrule
Correlation & 11\% & 8\% & 3\% & 10\% & 0.64 $\to$ 0.69 & 0.15 $\to$ 0.16 & 1.1 $\to$ 1.2 \\
Granger & 13\% & 10\% & 4\% & 11\% & 0.79 $\to$ 0.83 & 0.21 $\to$ 0.24 & 1.7 $\to$ 1.7 \\
VARLiNGAM & 16\% & 12\% & 4\% & 12\% & 0.66 $\to$ 0.70 & 0.30 $\to$ 0.34 & 2.3 $\to$ 2.4 \\
DYNOTEARS & \multicolumn{4}{c}{2 wrong selections in all} & 0.28 $\to$ 0.28 & 1.00 $\to$ 1.00 & 7.3 $\to$ 6.6 \\
PC & 17\% & 12\% & 5\% & 12\% & 0.25 $\to$ 0.29 & 0.30 $\to$ 0.35 & 2.3 $\to$ 2.5 \\
PCMCI+ & 32\% & 19\% & 13\% & 11\% & 0.78 $\to$ 0.79 & 0.69 $\to$ 0.75 & 5.3 $\to$ 5.3 \\
SpaceTime & 15\% & 14\% & 1\% & 4\% & 0.63 $\to$ 0.65 & 0.74 $\to$ 0.76 & 5.9 $\to$ 5.5 \\
DoubleML & 40\% & 31\% & 9\% & 14\% & 0.67 $\to$ 0.67 & 0.72 $\to$ 0.80 & 5.5 $\to$ 5.5 \\
\bottomrule
\end{tabular}

\end{table}

\section{Additional results}
\label{app:results}
\label{app:errors}
\label{app:tables}
\label{app:configs}
\label{app:threshold}

\paragraph{Errors by role.} Table~\ref{tab:errors} breaks the wrong selections of each method down by the role of the
selected candidate, over all plants of all configurations, and compares each share with the share of that role among
all wrong candidates. Section~\ref{sec:experiments} reads the table.

\begin{table}[h]
\centering\scriptsize
\setlength{\tabcolsep}{3pt}
\caption{Errors of each method over the \ErrTPlants{} plants of all \NCells{} configurations: the share
of each role among the method's wrong selections, against the share among all wrong candidates (\emph{random choice}).
A cell is shaded when the method picks that role more often than a random choice would, darker the more often.
\emph{Nothing selected}: share of plants on which the
method selects nothing. The disturbance of the controlled cause (Section~\ref{sec:coupling}) is an ordinary ancestor with the controller
off, as in all configurations where it is scored.}
\label{tab:errors}
\resizebox{\linewidth}{!}{\begin{tabular}{@{}l c cccc c@{}}
\toprule
 & & \multicolumn{4}{c}{share of the wrong selections} & \\
\cmidrule(lr){3-6}
Method & wrong per plant & descendants & \shortstack{disturbance of the\\controlled cause} & other ancestors & siblings & nothing selected \\
\midrule
random choice & & 13\% & 5\% & 31\% & 51\% & \\
\midrule
Correlation & 14.7 & \cellcolor{shadepurple!14}16\% & \cellcolor{shadepurple!10}5\% & 30\% & 48\% & \cellcolor{shadetan!6}0\% \\
Granger & 11.6 & 12\% & \cellcolor{shadepurple!15}6\% & \cellcolor{shadepurple!10}33\% & 49\% & \cellcolor{shadetan!11}10\% \\
VARLiNGAM & 5.7 & \cellcolor{shadepurple!24}23\% & \cellcolor{shadepurple!22}8\% & 28\% & 42\% & \cellcolor{shadetan!9}5\% \\
DYNOTEARS & 0.0 & \multicolumn{4}{c}{2 wrong selections in all} & \cellcolor{shadetan!45}72\% \\
PC & 3.2 & \cellcolor{shadepurple!49}55\% & 2\% & 18\% & 25\% & \cellcolor{shadetan!6}0\% \\
PCMCI+ & 0.8 & 12\% & \cellcolor{shadepurple!38}14\% & \cellcolor{shadepurple!16}40\% & 35\% & \cellcolor{shadetan!11}10\% \\
SpaceTime & 0.3 & 5\% & \cellcolor{shadepurple!46}18\% & \cellcolor{shadepurple!10}33\% & 44\% & \cellcolor{shadetan!14}16\% \\
DoubleML & 0.7 & 6\% & \cellcolor{shadepurple!59}28\% & \cellcolor{shadepurple!12}36\% & 30\% & \cellcolor{shadetan!21}27\% \\
\bottomrule
\end{tabular}
}
\end{table}

\begin{figure}[h]
\centering
\includegraphics[width=0.92\linewidth]{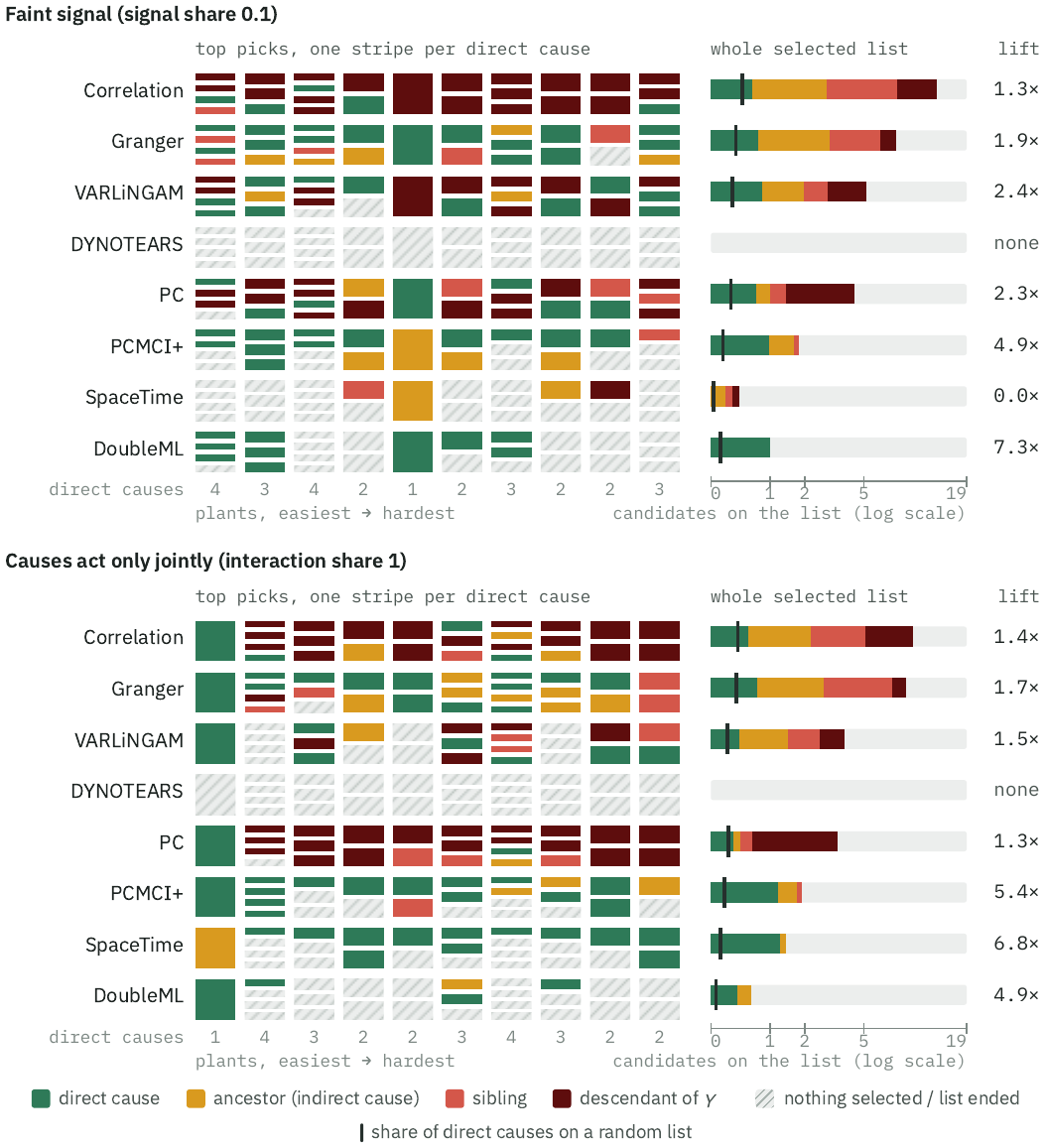}
\caption{The selected lists, as in Figure~\ref{fig:money}, for two single difficulties: a faint signal
(signal share 0.1, top) and causes that act only jointly (interaction share 1, bottom).}
\label{fig:money-appx}
\end{figure}

\paragraph{All configurations.} Figure~\ref{fig:money-appx} shows the selected lists for two single difficulties, as
Figure~\ref{fig:money} does for the reference and the combined difficulties. Table~\ref{tab:allresults} gives, for
every one of the \NCells{} configurations, the first pick, the mAP of the ranking and the precision of the selected
list of every method; the configurations are grouped by the difficulty they vary, with the symbols of
Sections~\ref{sec:shares} and~\ref{sec:optional}.

\begin{landscape}
\scriptsize
\setlength{\tabcolsep}{1pt}
\begin{longtable}{@{}l *{8}{c}@{\hspace{9pt}}*{8}{c}@{\hspace{9pt}}*{8}{c}@{}}
\caption{\small Every configuration and method, grouped by the share or component that the configuration changes: \emph{first pick}, plants (of 10) whose first selected candidate is a direct cause; \emph{mAP}, mean average precision of the ranking; \emph{precision}, of the selected list over the plants where the method selects something (-- when it never does). Each row departs from the reference configuration in the parameters named; \emph{combined difficulties}: $s_{\mathrm{sig}}$ 0.3, $s_{\mathrm{conf}} = s_{\mathrm{up}}$ 0.6, $s_{\mathrm{desc}}$ 0.5 and $s_{\mathrm{int}}$ 1 (Figure~\ref{fig:money}, bottom). Darker shading means a higher value.}\label{tab:allresults}\\
\toprule
 & \multicolumn{8}{c}{first pick} & \multicolumn{8}{c}{mAP} & \multicolumn{8}{c}{precision} \\
\cmidrule(lr){2-9}\cmidrule(lr){10-17}\cmidrule(l){18-25}
Configuration & \rotatebox{90}{Correlation} & \rotatebox{90}{Granger} & \rotatebox{90}{VARLiNGAM} & \rotatebox{90}{DYNOTEARS} & \rotatebox{90}{PC} & \rotatebox{90}{PCMCI+} & \rotatebox{90}{SpaceTime} & \rotatebox{90}{DoubleML} & \rotatebox{90}{Correlation} & \rotatebox{90}{Granger} & \rotatebox{90}{VARLiNGAM} & \rotatebox{90}{DYNOTEARS} & \rotatebox{90}{PC} & \rotatebox{90}{PCMCI+} & \rotatebox{90}{SpaceTime} & \rotatebox{90}{DoubleML} & \rotatebox{90}{Correlation} & \rotatebox{90}{Granger} & \rotatebox{90}{VARLiNGAM} & \rotatebox{90}{DYNOTEARS} & \rotatebox{90}{PC} & \rotatebox{90}{PCMCI+} & \rotatebox{90}{SpaceTime} & \rotatebox{90}{DoubleML} \\
\midrule
\endfirsthead
\toprule
 & \multicolumn{8}{c}{first pick} & \multicolumn{8}{c}{mAP} & \multicolumn{8}{c}{precision} \\
\cmidrule(lr){2-9}\cmidrule(lr){10-17}\cmidrule(l){18-25}
Configuration & \rotatebox{90}{Correlation} & \rotatebox{90}{Granger} & \rotatebox{90}{VARLiNGAM} & \rotatebox{90}{DYNOTEARS} & \rotatebox{90}{PC} & \rotatebox{90}{PCMCI+} & \rotatebox{90}{SpaceTime} & \rotatebox{90}{DoubleML} & \rotatebox{90}{Correlation} & \rotatebox{90}{Granger} & \rotatebox{90}{VARLiNGAM} & \rotatebox{90}{DYNOTEARS} & \rotatebox{90}{PC} & \rotatebox{90}{PCMCI+} & \rotatebox{90}{SpaceTime} & \rotatebox{90}{DoubleML} & \rotatebox{90}{Correlation} & \rotatebox{90}{Granger} & \rotatebox{90}{VARLiNGAM} & \rotatebox{90}{DYNOTEARS} & \rotatebox{90}{PC} & \rotatebox{90}{PCMCI+} & \rotatebox{90}{SpaceTime} & \rotatebox{90}{DoubleML} \\
\midrule
\endhead
\bottomrule
\endlastfoot
\multicolumn{25}{@{}l}{\textit{Reference} (5)} \\
\quad $s_{\mathrm{int}}$ 0 (reference) & \cellcolor{shade!60}10 & \cellcolor{shade!60}10 & \cellcolor{shade!60}10 & \cellcolor{shade!38}6 & \cellcolor{shade!22}3 & \cellcolor{shade!60}10 & \cellcolor{shade!49}8 & \cellcolor{shade!60}10 & \cellcolor{shade!49}0.79 & \cellcolor{shade!48}0.78 & \cellcolor{shade!51}0.84 & \cellcolor{shade!57}0.94 & \cellcolor{shade!23}0.31 & \cellcolor{shade!58}0.96 & \cellcolor{shade!48}0.77 & \cellcolor{shade!57}0.94 & \cellcolor{shade!14}0.14 & \cellcolor{shade!14}0.15 & \cellcolor{shade!24}0.33 & \cellcolor{shade!60}1.00 & \cellcolor{shade!25}0.35 & \cellcolor{shade!49}0.79 & \cellcolor{shade!52}0.85 & \cellcolor{shade!44}0.70 \\
\qquad 100 batches & \cellcolor{shade!55}9 & \cellcolor{shade!55}9 & \cellcolor{shade!60}10 & \cellcolor{shade!38}6 & \cellcolor{shade!17}2 & \cellcolor{shade!60}10 & \cellcolor{shade!49}8 & \cellcolor{shade!60}10 & \cellcolor{shade!51}0.83 & \cellcolor{shade!47}0.76 & \cellcolor{shade!53}0.87 & \cellcolor{shade!56}0.93 & \cellcolor{shade!23}0.31 & \cellcolor{shade!58}0.96 & \cellcolor{shade!44}0.71 & \cellcolor{shade!53}0.88 & \cellcolor{shade!14}0.15 & \cellcolor{shade!15}0.17 & \cellcolor{shade!20}0.27 & \cellcolor{shade!60}1.00 & \cellcolor{shade!26}0.37 & \cellcolor{shade!48}0.78 & \cellcolor{shade!57}0.94 & \cellcolor{shade!51}0.83 \\
\end{longtable}

\end{landscape}

\paragraph{Sensitivity to the selection threshold.} Every method selects a list from its ranking with a fixed rule:
correlation and Granger causality keep the candidates whose Benjamini--Hochberg adjusted $p$-value is below 0.05, and
DYNOTEARS those with a contemporaneous weight of at least $\tau_W = 0.3$ or a lagged weight of at least
$\tau_A = 0.1$, the thresholds of its paper \citep{pamfil2020dynotears}. Table~\ref{tab:selection} gives the results at
other thresholds; the ranking metrics do not depend on the threshold. The
long lists of correlation and Granger causality come from this threshold: keeping only their first three
candidates raises their precision from \Sel{correlation}{0}{prec} and \Sel{granger}{0}{prec} to
\Sel{correlation}{5}{prec} and \Sel{granger}{5}{prec}. DYNOTEARS selects nothing on \Sel{dynotears}{0}{none} of the
plants at its paper's thresholds; at $\tau_W = 0.03$, $\tau_A = 0.01$ it selects nothing on \Sel{dynotears}{2}{none},
its first pick is right on \Sel{dynotears}{2}{first} of the plants and its precision is \Sel{dynotears}{2}{prec}
(PCMCI+ at 0.05: \Ovr{pcmciplus}{first} and \Ovr{pcmciplus}{prec}). For PCMCI+ we report 0.05, the level of our
other tests; the
experiments of its paper use 0.01. A stricter level shortens its list and raises its precision, from
\Sel{pcmciplus}{1}{prec} at 0.05 to \Sel{pcmciplus}{3}{prec} at 0.001, but leaves it empty on
\Sel{pcmciplus}{3}{none} of the plants instead of \Sel{pcmciplus}{1}{none}, with fewer first picks right
(\Sel{pcmciplus}{3}{first} instead of \Sel{pcmciplus}{1}{first}).

\begin{table}[h]
\centering\small
\setlength{\tabcolsep}{3.5pt}
\caption{Selected lists under other thresholds, over all \SortPlants{} plants (columns as in Table~\ref{tab:overall}). BH:
Benjamini--Hochberg adjusted $p$-values; \emph{first 3 by score}: the three highest-ranked candidates.}
\label{tab:selection}
\begin{tabular}{@{}l l ccccc@{}}
\toprule
Method & Threshold & length & precision & lift & first pick & nothing selected \\
\midrule
Correlation & BH, $\alpha = 0.05$ (used) & 16.9 & 0.15 & 1.1 & 64\% & 0\% \\
 & BH, $\alpha = 10^{-3}$ & 13.2 & 0.17 & 1.3 & 64\% & 2\% \\
 & BH, $\alpha = 10^{-6}$ & 9.6 & 0.22 & 1.7 & 63\% & 6\% \\
 & BH, $\alpha = 10^{-10}$ & 7.1 & 0.26 & 2.0 & 61\% & 11\% \\
 & Bonferroni, $\alpha = 0.05$ & 14.2 & 0.16 & 1.3 & 64\% & 0\% \\
 & first 3 by score & 3.0 & 0.42 & 3.3 & 64\% & 0\% \\
\midrule
Granger & BH, $\alpha = 0.05$ (used) & 13.7 & 0.21 & 1.7 & 79\% & 10\% \\
 & BH, $\alpha = 10^{-3}$ & 9.9 & 0.29 & 2.3 & 74\% & 18\% \\
 & BH, $\alpha = 10^{-6}$ & 6.3 & 0.38 & 3.0 & 67\% & 27\% \\
 & BH, $\alpha = 10^{-10}$ & 4.0 & 0.49 & 3.9 & 61\% & 34\% \\
 & Bonferroni, $\alpha = 0.05$ & 11.0 & 0.26 & 2.0 & 79\% & 10\% \\
 & first 3 by score & 3.0 & 0.49 & 3.8 & 82\% & 0\% \\
\midrule
DYNOTEARS & $\tau_W = 0.3$, $\tau_A = 0.1$ (used) & 0.3 & 1.00 & 7.3 & 28\% & 71\% \\
 & $\tau_W = 0.1$, $\tau_A = 0.033$ & 1.5 & 0.89 & 6.9 & 71\% & 25\% \\
 & $\tau_W = 0.03$, $\tau_A = 0.01$ & 2.8 & 0.73 & 5.7 & 85\% & 8\% \\
 & $\tau_W = \tau_A = 0.01$ & 3.2 & 0.68 & 5.3 & 87\% & 5\% \\
\midrule
PCMCI+ & $\alpha = 0.2$ & 4.0 & 0.48 & 3.8 & 82\% & 2\% \\
 & $\alpha = 0.05$ (used) & 2.4 & 0.69 & 5.3 & 78\% & 10\% \\
 & $\alpha = 0.01$ & 1.8 & 0.80 & 6.1 & 73\% & 20\% \\
 & $\alpha = 0.001$ & 1.4 & 0.89 & 6.8 & 66\% & 30\% \\
\bottomrule
\end{tabular}

\end{table}

\clearpage
\section{Difficulty measured on the data}
\label{app:map}
\begin{figure}[!h]
\centering
\includegraphics[width=\linewidth]{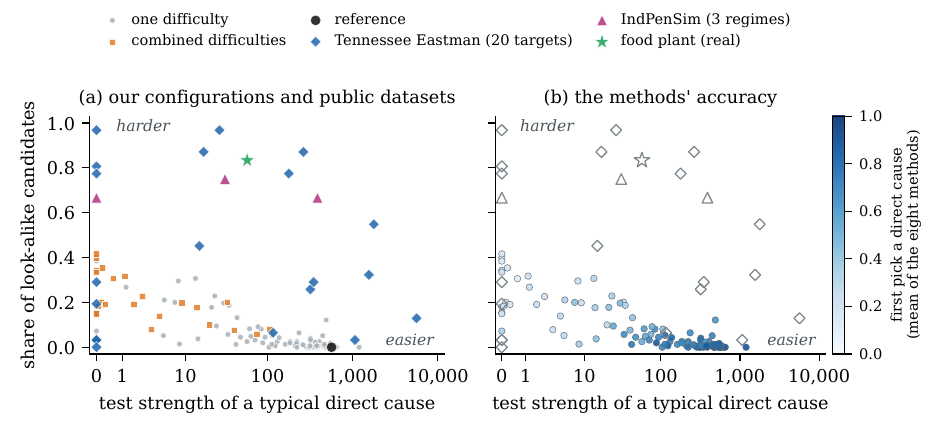}
\caption{Difficulty measured on the data. Horizontal: test strength of the median direct cause (logarithmic scale).
Vertical: share of wrong candidates that are look-alikes. (a) Our configurations and three public process datasets.
(b) Share of plants whose first pick is a direct cause, averaged over the eight methods; the public datasets are not
evaluated (hollow).}
\label{fig:map}
\end{figure}

Each configuration sets its difficulty through the shares. Figure~\ref{fig:map} measures the resulting difficulty on
the generated data with two quantities per plant, computed at the resolution and look-back the methods use: the test
strength of the direct causes and the share of wrong candidates that are look-alikes.

\emph{Test strength.} For each direct cause, its partial $R^2$ given the past of $\Y$ and the other direct causes, from
out-of-sample fits across batches, as the effect size $f^2 = R^2/(1-R^2)$, times the number of observations: the
expected statistic of a test for that cause. The figure shows the median over a plant's direct causes, so one faint
cause does not set the value.

\emph{Look-alikes.} A wrong candidate is a look-alike if its past carries more information about $\Y$ than the past
of the median direct cause, measured by $k$-nearest-neighbour mutual information. The definition uses only the
direct causes, so it applies to datasets whose other roles are unknown; in our plants, look-alikes are siblings,
ancestors and descendants. The vertical axis is the share of wrong candidates that are look-alikes.

Each configuration is summarised over its 10 plants. The public problems are Tennessee Eastman
\citep{downs1993te}, with the fault runs as episodes and every variable with a single parent in the CIPCaD graph
\citep{menegozzo2022cipcad} as a target; IndPenSim \citep{goldrick2019indpensim}, with the production rate as target
and one problem per control regime; and the food-processing line of CIPCaD-Bench, with the quality variable as target
and the last stage's variables as its direct causes.

The combined configurations lie at lower test strength and with more look-alikes than those with one difficulty, and
the methods' mean accuracy falls in the same direction (Figure~\ref{fig:map}b). The public datasets lie within the
range of test strength of our configurations, with more look-alikes than most of them. Both quantities summarise
difficulty for the target only. A component aimed at one assumption of a method, such as downstream measurements for
methods that ignore time order, can change the results of those methods with little change in either quantity; the
results are therefore reported per method.

\section{Method settings and reproducibility}
\label{app:methods}
\label{app:repro}
Lagged correlation ranks the candidates by their correlation with $\Y$ over lags; Granger causality tests whether a
candidate's past improves the prediction of $\Y$ beyond its own past. PC tests conditional independence between
values at the same time step; PCMCI+ tests lagged and contemporaneous links conditional on the estimated parents of
both variables. DYNOTEARS fits a structural vector autoregression under a continuous acyclicity constraint; VARLiNGAM
orients effects within a time step through non-Gaussian noise; SpaceTime scores graphs by minimum description length
with every batch as a context. DoubleML estimates the effect of a candidate's past on $\Y$ with flexible models of
all other candidates and selects the candidates whose effect is significant. DYNOTEARS uses the edge thresholds of
its paper (0.3 for contemporaneous and 0.1 for lagged edges).

The common look-back serves as Granger's maximum lag, PCMCI+'s $\tau_{\max}$, VARLiNGAM's lag order and the
window of lagged correlation, whose score is the largest absolute correlation over lags $0$ to $L$, tested with
Fisher's $z$-transform on $n - L$ effective samples.
PCMCI+ uses partial correlation, at most 5 conditioning variables and only links into the target. PC runs on the series at the evaluation resolution, without lags, at a significance level of 0.05 and in its
original variant with Fisher's $z$ test; a candidate is selected when PC links it to $\Y$ in either direction or
leaves the edge undirected. DYNOTEARS runs on standardised series, with its lag window set to the common look-back, regularisation
$\lambda_W = \lambda_A = 0.05$ and at most 100 iterations. VARLiNGAM runs on the batches concatenated in order; its list is the
candidates that keep a non-zero coefficient into $\Y$ after the adaptive-lasso pruning of the \texttt{lingam} library, and its ranking the
largest absolute coefficient over lags.

DoubleML fits one partially linear model per candidate with the partialling-out score. The data form a table: each
row is one step of a batch at the evaluation resolution, from step $L+1$ on, and lags never cross a batch (2,400
rows at the reference). The outcome is $\Y(t)$. The treatments are the candidate at lags $1,\dots,L$, and each
lag's coefficient is estimated with the other lags among the controls. The controls are $\Y$ at lags $1,\dots,L$
and every other recorded variable at lags $0,\dots,L$; the lag-0 values are included because one averaged step
contains much of a cause's effect. The nuisance models are two LightGBM regressors \citep{ke2017lightgbm} with
fixed parameters: 300 trees, learning rate 0.05, 31 leaves, at least 20 samples per leaf, and row and column
subsampling of 0.8. Cross-fitting uses three folds and one repetition. Folds contain whole batches and standard
errors are clustered by batch, which is how autocorrelation within a batch is handled. The $p$-values are adjusted
with Holm over a candidate's lags, the smallest is kept, and Benjamini--Hochberg at 0.05 runs across variables. The
ranking is the largest $|t|$ over lags.

SpaceTime scores graphs with Gaussian-process mechanisms. Every batch is a separate context, in which the
mechanisms may differ, and each batch has a single regime, with no change of mechanism over time. The search is
restricted to the parents of $\Y$, as the task asks, and $\Y$ is constrained to have no children, so that every edge
found at $\Y$ points into it. Its list is the parents it names at its default acceptance level. SpaceTime returns a set rather than a ranking, so we rank the
parents it names first and the other candidates by their pairwise gain, the improvement in description length when
each alone is a parent of $\Y$, which its search computes in its first step.

The sortability checks rank the candidates in ascending order of their variance or of the $R^2$ of a linear
regression on the other candidates, so the least predictable candidate comes first. All random seeds are fixed.
Every method sees the same table: the block means of every batch at the evaluation resolution. Lags are constructed
within batches, except for VARLiNGAM, which uses the concatenated batches.

\end{document}